\documentclass[journal]{IEEEtran}

\usepackage{amsmath,graphicx,mathrsfs}
\usepackage{bbm}
\usepackage{etex}
\usepackage{float,subfigure}
 
\usepackage{color}

\usepackage{comment}
\usepackage{hyperref}

\usepackage{fontawesome}
\usepackage{cite}
\usepackage{amsmath,amssymb,amsfonts}
\usepackage{algorithmic}
\usepackage{graphicx}
\usepackage{textcomp}
\usepackage{booktabs}
\usepackage{multirow}
\usepackage{footnote}
\usepackage{stfloats}
\usepackage{pifont}
\usepackage[dvipsnames,table]{xcolor}
\usepackage{arydshln} 
\usepackage{hyperref}
\colorlet{lightgray}{gray!10}
\colorlet{deepgray}{gray!130}
\colorlet{lightred}{red!50}
\hypersetup{
	colorlinks=true,
	linkcolor=red,
	filecolor=blue,      
	urlcolor=blue,
	citecolor=blue,
}
\ifCLASSINFOpdf
\else
\fi

\begin{document}
	
	\title{No-Reference Image Quality Assessment with Global-Local Progressive Integration and Semantic-Aligned Quality Transfer}
	\author{Xiaoqi Wang,  Yun Zhang,~\IEEEmembership{Senior Member,~IEEE} 
 
		\thanks{Xiaoqi Wang and Yun Zhang are with the School of Electronics and Communication Engineering, Sun Yat-sen University, Shenzhen, 518107, China (e-mail: \href{mailto:wangxq79@mail2.sysu.edu.cn}{\textcolor{black}{wangxq79@mail2.sysu.edu.cn}}; \href{mailto:zhangyun2@mail.sysu.edu.cn}{\textcolor{black}{zhangyun2@mail.sysu.edu.cn}}).} 
	 
	}
	% The paper headers
	\markboth{~Vol.~XX, No.~XX,	February ~2025}
	{Shell \MakeLowercase{\textit{et al.}}: Bare Demo of IEEEtran.cls for IEEE Journals}
	
	\maketitle

\begin{abstract}
Accurate measurement of image quality without reference signals remains a fundamental challenge in low-level visual perception applications. In this paper, we propose a global-local progressive integration model that addresses this challenge through three key contributions: 1)~We develop a dual-measurement framework that combines vision Transformer (ViT)-based global feature extractor and convolutional neural networks (CNNs)-based local feature extractor to comprehensively capture and quantify image distortion characteristics at different granularities. 2)~We propose a progressive feature integration scheme that utilizes multi-scale kernel configurations to align global and local features, and progressively aggregates them via an interactive stack of channel-wise self-attention and spatial interaction modules for multi-grained quality-aware representations. 3)~We introduce a semantic-aligned quality transfer method that extends the training data by automatically labeling the quality scores of diverse image content with subjective opinion scores. Experimental results demonstrate that our model yields 5.04\% and 5.40\% improvements in Spearman's rank-order correlation coefficient (SROCC) for cross-authentic and cross-synthetic dataset generalization tests, respectively. Furthermore, the proposed semantic-aligned quality transfer further yields 2.26\% and 13.23\% performance gains in evaluations on single-synthetic and cross-synthetic datasets\footnote{The codes and proposed dataset will be released  at~\href{https://github.com/XiaoqiWang/GlintIQA}{\color{VioletRed}{here.}}}.
\end{abstract}

\begin{IEEEkeywords}
	Image quality assessment, dataset construction, label transfer, feature integration, vision transformer.
\end{IEEEkeywords}

	\IEEEpeerreviewmaketitle

\section{Introduction} 
\IEEEPARstart{A}{s} a crucial measurement component in modern imaging systems, objective image quality assessment (IQA) enables quality control and optimization across various applications including medical imaging, visual inspection, and multimedia services. IQA aims to enable automated quantitative evaluation of visual image quality in alignment with human perception. Based on the availability of reference signals, IQA methods are categorized into three paradigms: full-reference (FR-IQA), reduced-reference (RR-IQA), and no-reference IQA (NR-IQA)~\cite{iqasurvey}. Among these, NR-IQA has gained significant research attention due to its practical applicability in real-world scenarios where reference images are unavailable.

Traditional NR-IQA methods~\cite{biqi, diivine, ilniqe, brisque,Yang_tim2022} extract predefined measurement features, \textit{\textit{e.g.}}, wavelet coefficients~\cite{biqi} and normalized luminance statistics~\cite{brisque}, which are then calibrated using regression models like support vector regression~(SVR)~\cite{diivine} and multivariate Gaussian models~(MVG) \cite{ilniqe}. However, these methods face limitations in measurement accuracy and reliability due to their dependence on manually designed features and inability to comprehensively characterize complex image degradations. In contrast, deep learning-based NR-IQA models~\cite{ meon,rankiqa, WaDIQaM, diqa,Jiang_tim2020, dbcnn,hyperiqa,SAWAR} demonstrate superior performance by end-to-end learning quality-relevant features from training data.

The effectiveness of deep learning-based NR-IQA methods depends on two main components:~measurement model and training data. Convolutional neural networks (CNNs) serve as effective tools for feature extraction and quality assessment. Through spatially-sliding convolutional kernels, the networks analyze local image regions to detect visual patterns and transform them into discriminative quality-relevant features. Early CNNs-based NR-IQA models\cite{diqa,WaDIQaM} primarily employed single-stream architectures, stacking CNN layers and linear layers to regress image quality scores in an end-to-end manner. To enhance the assessment accuracy, researchers have developed increasingly sophisticated architectures~\cite{twostream, meon, dbcnn, DACNN, rankiqa, aigqa, cahdc}. Wang et al.\cite{twostream} integrated distorted images with their corresponding gradient maps to augment the extraction of structural detail features. Ma et al.\cite{meon} introduced an auxiliary distortion recognition branch into the quality prediction network, accounting for the impact of diverse distortion types on image quality perception. Zhang et al.\cite{dbcnn} and Pan et al.\cite{DACNN} presented a dual-stream architecture that integrates semantic and distortion perception, enabling adaptation to both authentic and synthetic distortion measurement scenarios. Liu et al.\cite{rankiqa} employed Siamese networks to establish pairwise rankings among images exhibiting varying levels of distortion for learning relative quality representations. Further advancing architectural complexity, Wu et al.\cite{aigqa} proposed a multi-stream structure that simultaneously extracts features from distorted images, reconstructed images, error maps, and structural degradation maps. Wu et al.\cite{cahdc} adopted hierarchical structures that employ modules across different stages of the model to capture features across diverse scales. 

However, despite these advancements, CNNs' inherent local processing nature\cite{visualizing} can limit their capacity to model global dependencies within images. This limitation has driven researchers to explore the use of vision Transformers (ViTs) that have shown exceptional ability in capturing long-range spatial correlations. The integration of ViTs into NR-IQA has followed two distinct approaches. The first approach uses ViTs primarily as regression modules while retaining CNNs for feature extraction\cite{triq, tres, sgtnet}. You et al.\cite{triq} leveraged Transformer to establish a mapping from semantic features to quality scores. Golestaneh et al.\cite{tres} further constrained the features to represent quality consistency under flipping and learned the relative quality ordering among images. Zhu et al.\cite{sgtnet} introduced a saliency map within the feature space to guide the Transformer's attention towards regions of interest during regression. The second approach employs ViTs as the primary feature extraction backbone\cite{mstriq,deiqt}. Wang et al.\cite{mstriq} built a swin Transformer-based siamese network trained jointly on regression and pairwise ranking tasks. Qin et al.\cite{deiqt} proposed an attention mechanism on transformer-encoded features, mimicking diverse human evaluations for quality prediction from various perspectives. However, both approaches face challenges in NR-IQA applications. The first approach constrains feature representation by limiting ViTs to post-processing CNNs-extracted features~\cite{tres, triq}. The second approach encounters difficulties due to ViTs' patch-based tokenization~\cite{vit,convit}, where each token represents a low-resolution image patch, potentially compromises the measurement of spatial details critical for quality evaluation.

Moreover, while ViTs excel at capturing global dependencies, their lack of inductive biases necessitates larger datasets to achieve comparable performance~\cite{coatnet}. However, commonly used synthetic datasets~\cite{live,tid2013,kadid-10k} often lack content diversity. For instance, TID2013~\cite{tid2013} contains only 3,000 images across 25 content types, and KADID-10k~\cite{kadid-10k} provides 10,125 images with 81 content types. Expanding these datasets is constrained by the high costs and complexities of subjective quality evaluation procedures, which impacts ViTs' effectiveness in understanding diverse image degradations. Researchers have attempted to address the data limitations through various augmentation techniques. The patch-based approach randomly selects small patches (\textit{e.g.}, 112$\times$112 as used in \cite{diqa}) from the full images. To mitigate potential label bias from small patches, Kim et al.\cite{biecon} utilized patch-level quality scores from FR-IQA methods as intermediate regression targets. Several studies\cite{meon,rankiqa,aigqa,cahdc} have created large-scale datasets for pre-training NR-IQA models. Ma et al.\cite{meon} synthesized images with different distortions using artificial degradation for training distortion classification networks. Liu et al.\cite{rankiqa} leveraged injected distortion levels of synthetic images as quality rankings. Previous attempts at data augmentation, including distortion level injection\cite{rankiqa} or FR-IQA algorithm annotation\cite{cahdc}, only provide coarse quality indicators or show distortion-specific biases\cite{liu_tip2013}. While FR-IQA algorithms can generate fine-grained image quality scores, they often exhibit biases towards certain distortion types and rely primarily on low-level feature discrepancies, overlooking the crucial role of high-level semantic content in visual quality perception.

Based on the issues stated above, this paper focuses on two critical aspects: model architecture and training data. Our main contributions are three-fold:
\begin{itemize}
	\item To overcome architectural limitations, we propose a \textbf{G}lobal-\textbf{L}ocal progressive \textbf{Int}egration model (termed GlintIQA) for image quality assessment. This architecture integrates ViT-based global feature extraction (VGFE) with CNN-based local feature extraction (CLFE), leveraging ViTs' strength in capturing long-range dependencies through self-attention mechanisms while preserving CNNs' capability in quantifying fine-grained spatial details. This dual-stream design effectively overcomes both the patch-level information loss and inductive bias constraints in pure ViT-based approaches and the local context constraints in CNN-only architectures.
	
	\item To facilitate effective feature integration, we develop an efficient progressive integration strategy that considers both channel-wise and spatial information across features of different granularities. Our approach introduces cascaded channel-wise self-attention (CWSA) modules that dynamically adjust feature importance across different granularities, coupled with spatial interaction enhancement modules (SIEM) that strengthen the interactions between cross-scale local and global features, leading to improved quality assessment accuracy.

	\item To address the challenges of data scarcity and annotation limitations in model training, we propose a semantic-aligned quality transfer~(SAQT) method for automated image quality annotation. This approach exploits perceptual correlations between semantically similar images under identical distortion conditions to transfer subjective quality scores. Through statistical validation, we demonstrate the efficacy of the SAQT-based annotation method and establish the SAQT-IQA dataset, which yields substantial improvements in model generalization capabilities when employed for training.
\end{itemize}
 
The paper is structured as follows:  
Section \ref{dataset_construction} describes the quality labeling method and the construction of the proposed SAQT-IQA dataset. Section \ref{methodology} details the proposed GlintIQA model. Section \ref{experiment} presents comprehensive experimental results and ablation studies across multiple IQA benchmark databases. Finally, Section \ref{conclusion} draws a conclusion.

\begin{figure*}[t!]
	\centering
	\subfigure[CSIQ~\cite{csiq}]{%
		\begin{minipage}[t]{0.23\linewidth}
			\centering
			\includegraphics[width=\linewidth]{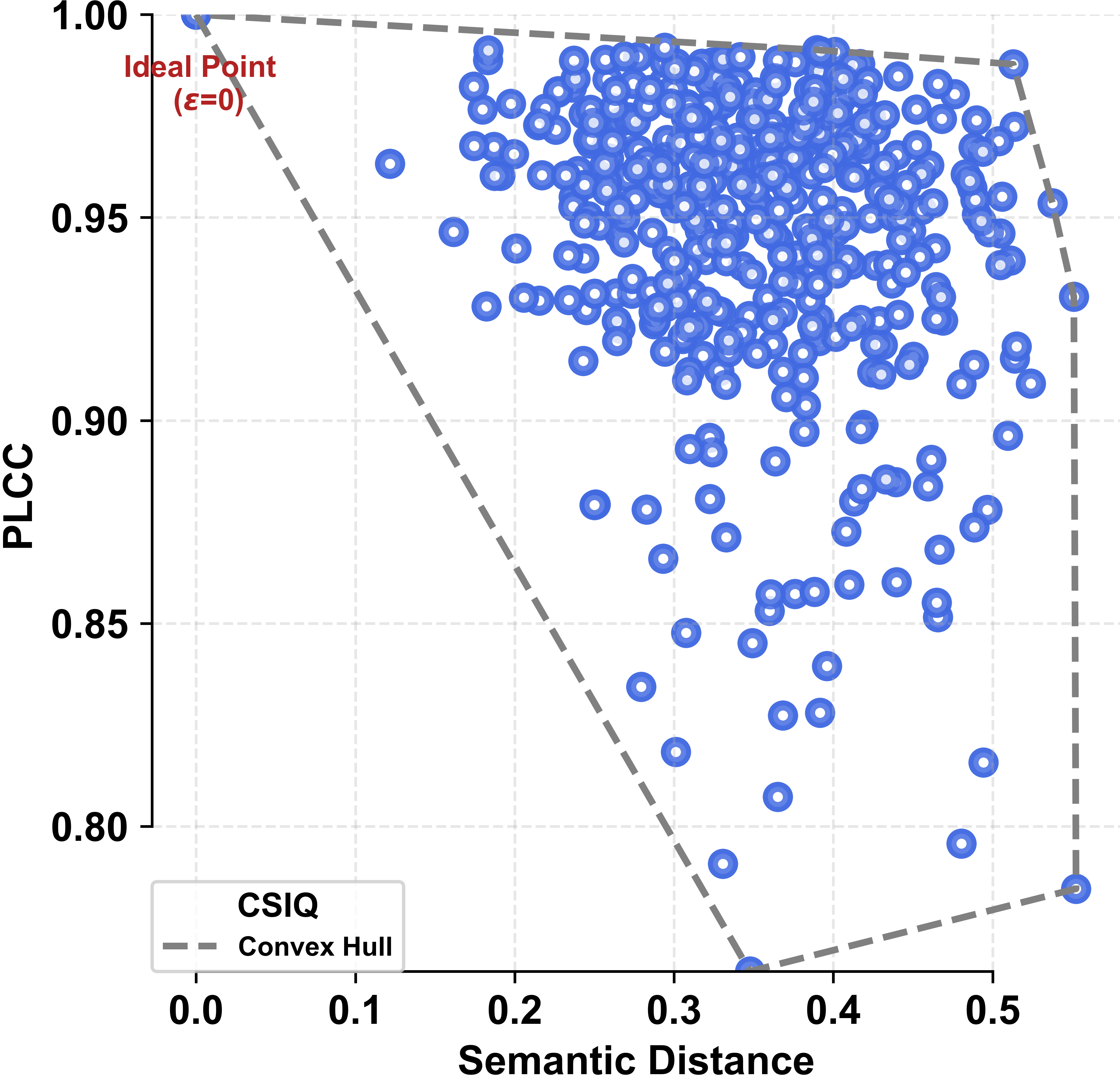}\noindent\vspace{-1mm}
			%			\label{fig:correlation}
		\end{minipage}%
	}%
	\subfigure[TID2013~\cite{tid2013}]{%
		\begin{minipage}[t]{0.23\linewidth}
			\centering
			\includegraphics[width=\linewidth]{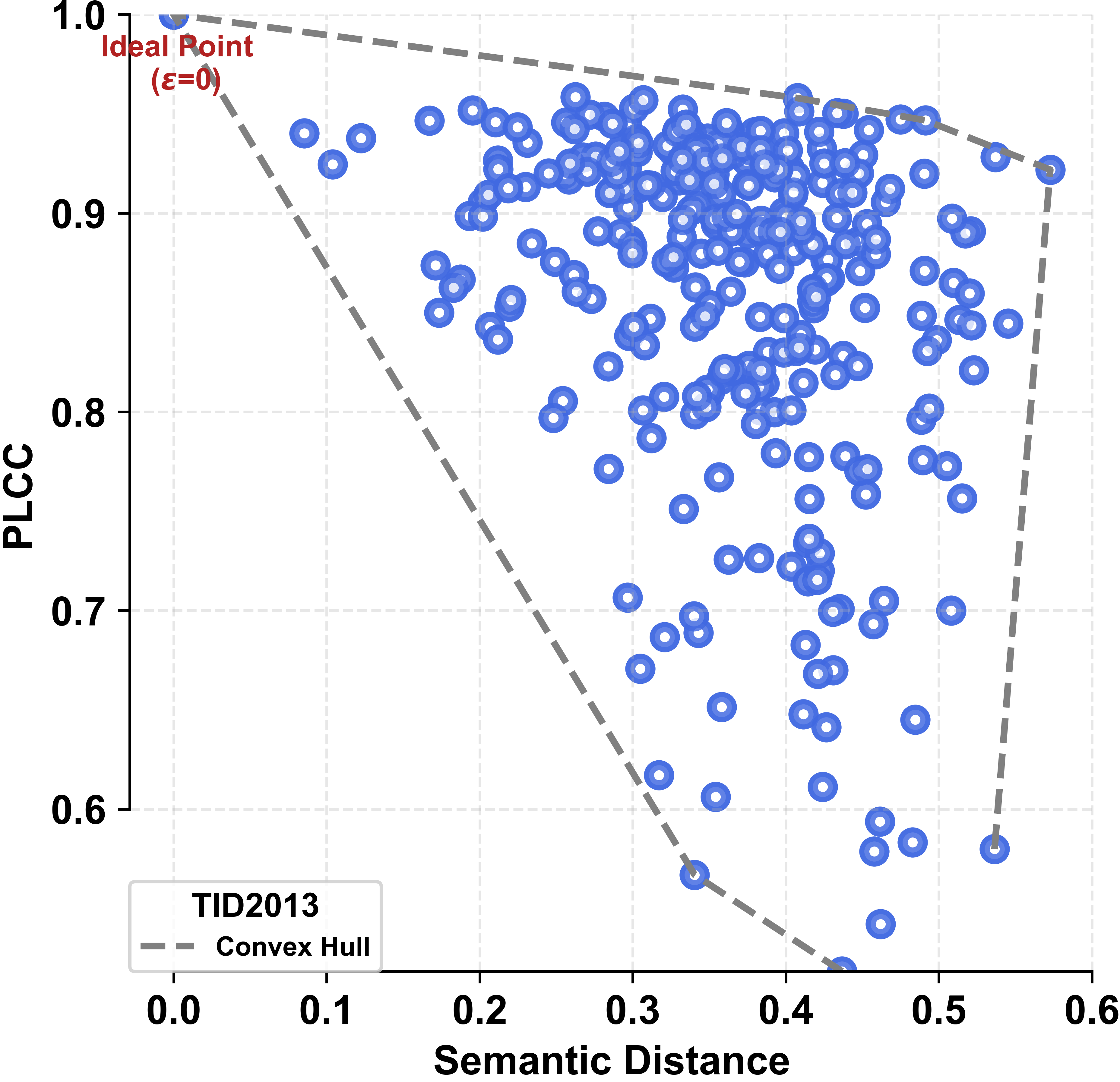}\noindent\vspace{-1mm}
			%			\label{fig:accuracy}c
		\end{minipage}
	}
	\subfigure[KADID-10k~\cite{kadid-10k}]{%
		\begin{minipage}[t]{0.23\linewidth}
			\centering
			\includegraphics[width=\linewidth]{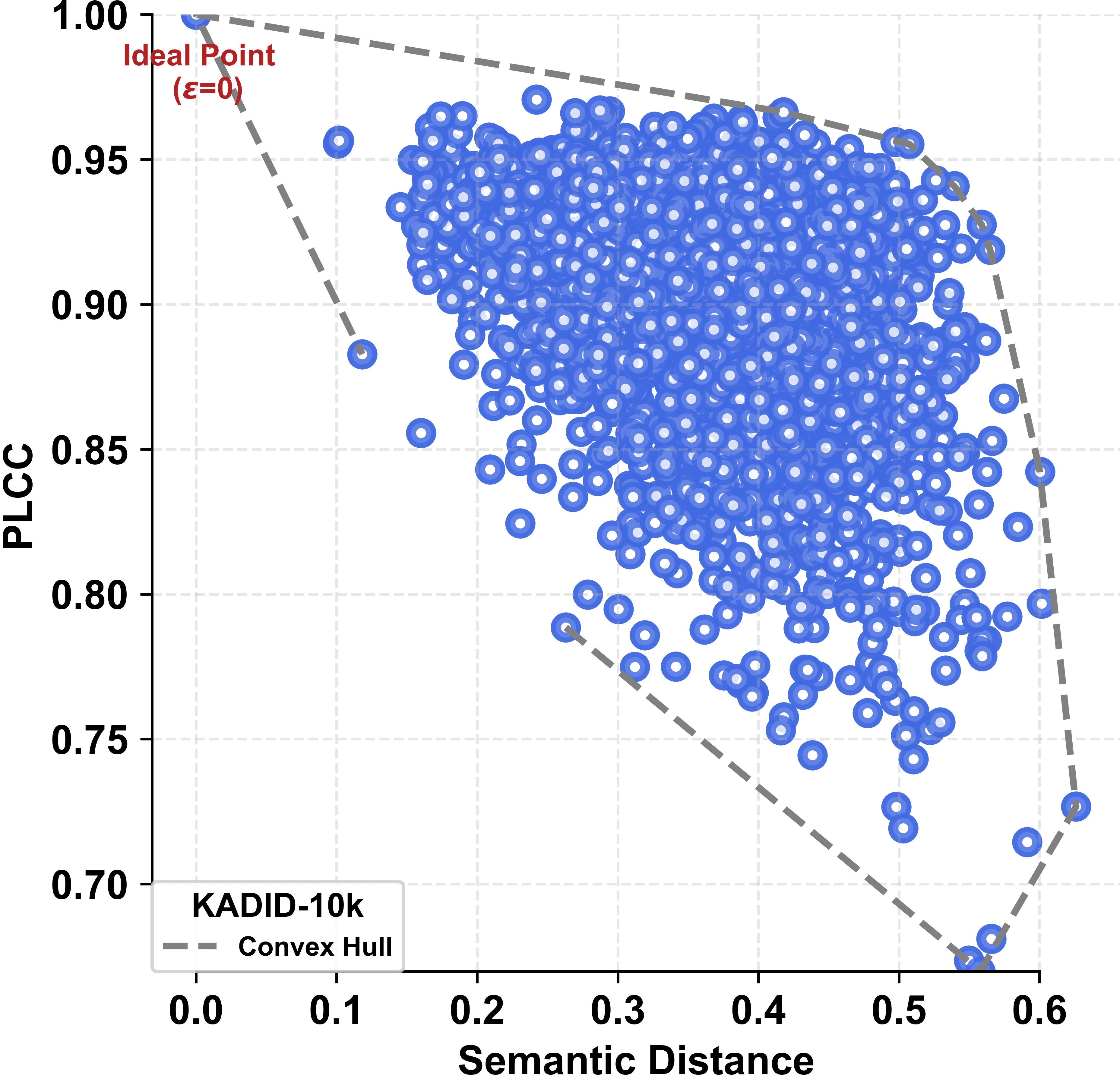}\noindent\vspace{-1mm}
			%			\label{fig:correlation}
		\end{minipage} \label{fig:analysis_c}
	}%
	\subfigure[LIVE-MD~\cite{livemd}]{%
		\begin{minipage}[t]{0.23\linewidth}
			\centering
			\includegraphics[width=\linewidth]{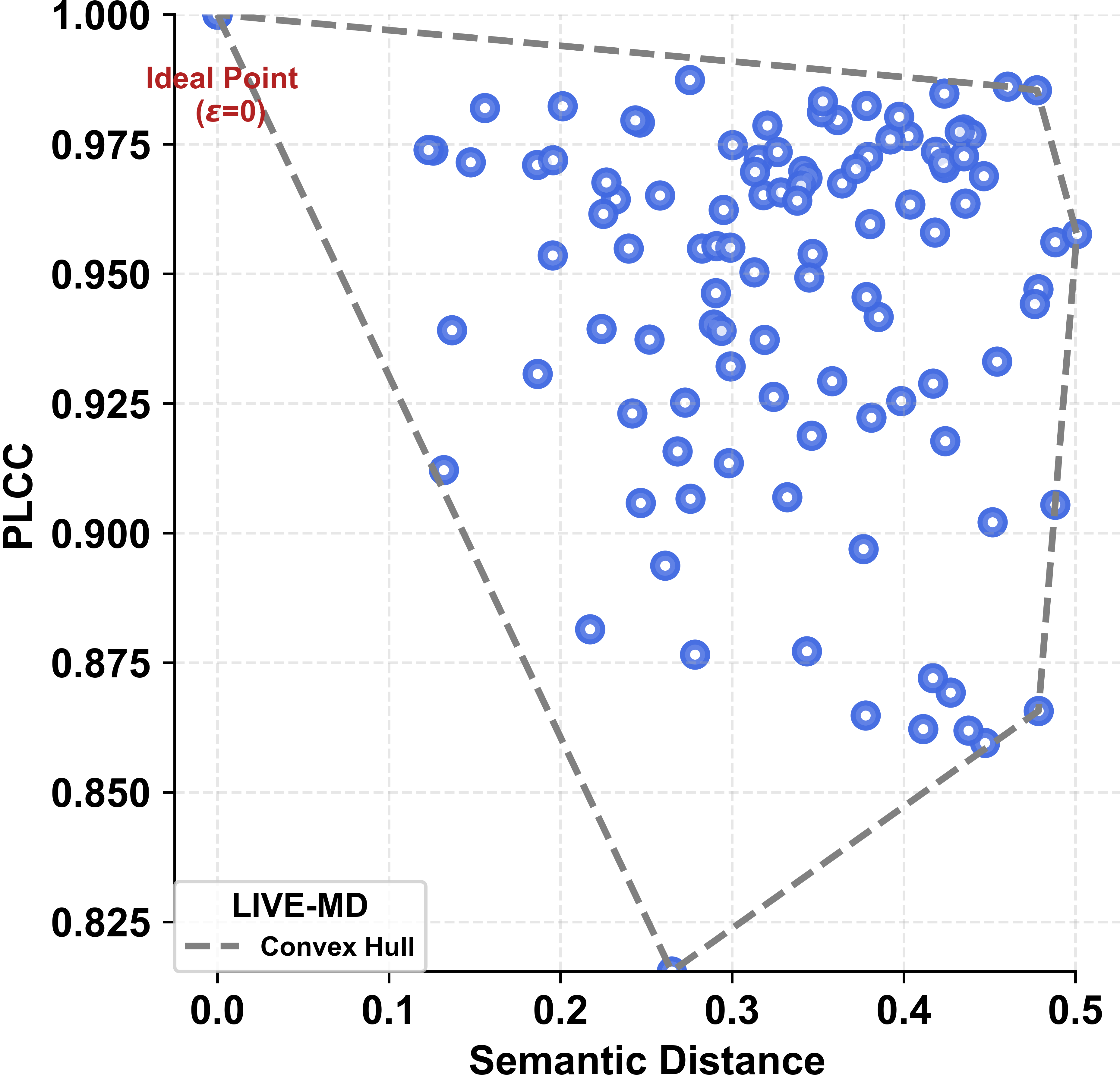}\noindent\vspace{-1mm}
			%			\label{fig:accuracy}c
		\end{minipage}
	}
	\caption{The correlation between image semantic distance and quality scores is demonstrated under identical distortion conditions across four IQA datasets.}
	\label{fig:analysis}
\end{figure*}

\section{The proposed semantic-aligned quality transfer method} \label{dataset_construction} 
We propose semantic-aligned quality transfer method that leverages the perceptual correlation of similar content under identical distortions~\cite{instanceiqa} for labeling image quality score. We begin with a statistical analysis to validate the feasibility of our method (section \ref{sec:labeling_val}). Then, we detail the construction of a large-scale synthetic dataset with this method (section \ref{sec:labeling_dataset}).

\subsection{Statistcal Analysis}  \label{sec:labeling_val}
%\subsection{Empirical Validation}  \label{sec:labeling_val}
To validate the relationship between semantic similarity and quality perception, we conducted a comprehensive statistical analysis across four IQA benchmark datasets. These datasets consist of pristine images and their corresponding distorted versions, each associated with mean opinion scores (MOS). Fig.~\ref{fig:analysis} illustrates the relationship between semantic distance (on the x-axis) and the correlation of quality scores (on the y-axis) across these datasets. Specifically, the x-axis represents the cosine distance between semantic feature vectors (extracted by ResNet50) of pristine image pairs, while the y-axis indicates the correlation of quality scores for the distorted images corresponding to these pristine pairs. The results reveal that for image pairs with larger semantic distances, the correlation between their quality scores exhibits greater variability. In contrast, as the semantic distance between images decreases, the correlations become increased and more consistent, approaching a value of 1 as the distance nears zero. This trend is consistently observed across all four datasets, which reinforces the conclusion that images with similar content tend to exhibit comparable visual quality labels when subjected to identical distortion conditions. Building on these insights, the proposed labeling method leverages human-annotated images to transfer quality labels to degraded images with similar content, enabling the construction of a large-scale synthetic dataset guided by semantic similarity. Next, we describe the specific steps for constructing the expanded dataset.

%%%%%%%%%%%%%%%%%%%%%%%%%%%%%%%%%

\subsection{Dataset Construction} \label{sec:labeling_dataset}
\subsubsection{Preliminary Data Preparation} 
To create the dataset, we meticulously selected 50,000 high-quality images from the KADIS-700k~\cite{kadid-10k} and applied 25 types of distortions with five intensity levels to each image. The distortion types were selected to be consistent with those present in KADID-10k~\cite{kadid-10k}. Finally, this process produced 6,250,000 degraded images. The quality of the generated images is annotated using MOS derived from the KADID-10k IQA database. Let $ \mathcal{X} = \left\{(\textbf{X}_{i,d},\textbf{M}_{i,d})\right\}_{i=1}^P$ denote the set of pristine images and their corresponding degraded versions, where $\textbf{X}_{i,d}=\left\{x^{t,l}_{i,d}|t\in \mathcal{T}, l \in  \mathcal{L}\right\}$ represents the degraded images of the $i$-th pristine image, and $\textbf{M}_{i,d}=\left\{m^{t,l}_{i,d}|t\in  \mathcal{T}, l \in  \mathcal{L}\right\}$ are their corresponding MOS values, scaled to a range between 0 and 9. The high-quality images are represented by the set $\textbf{Y} = \left\{Y_j\right\}_{j=1}^H$, where $Y_j$ refers to the $j$-th high-quality image and $H$ is the total number of high-quality images.

\begin{figure}[!tp]
	\centering
	\includegraphics[scale=.55]{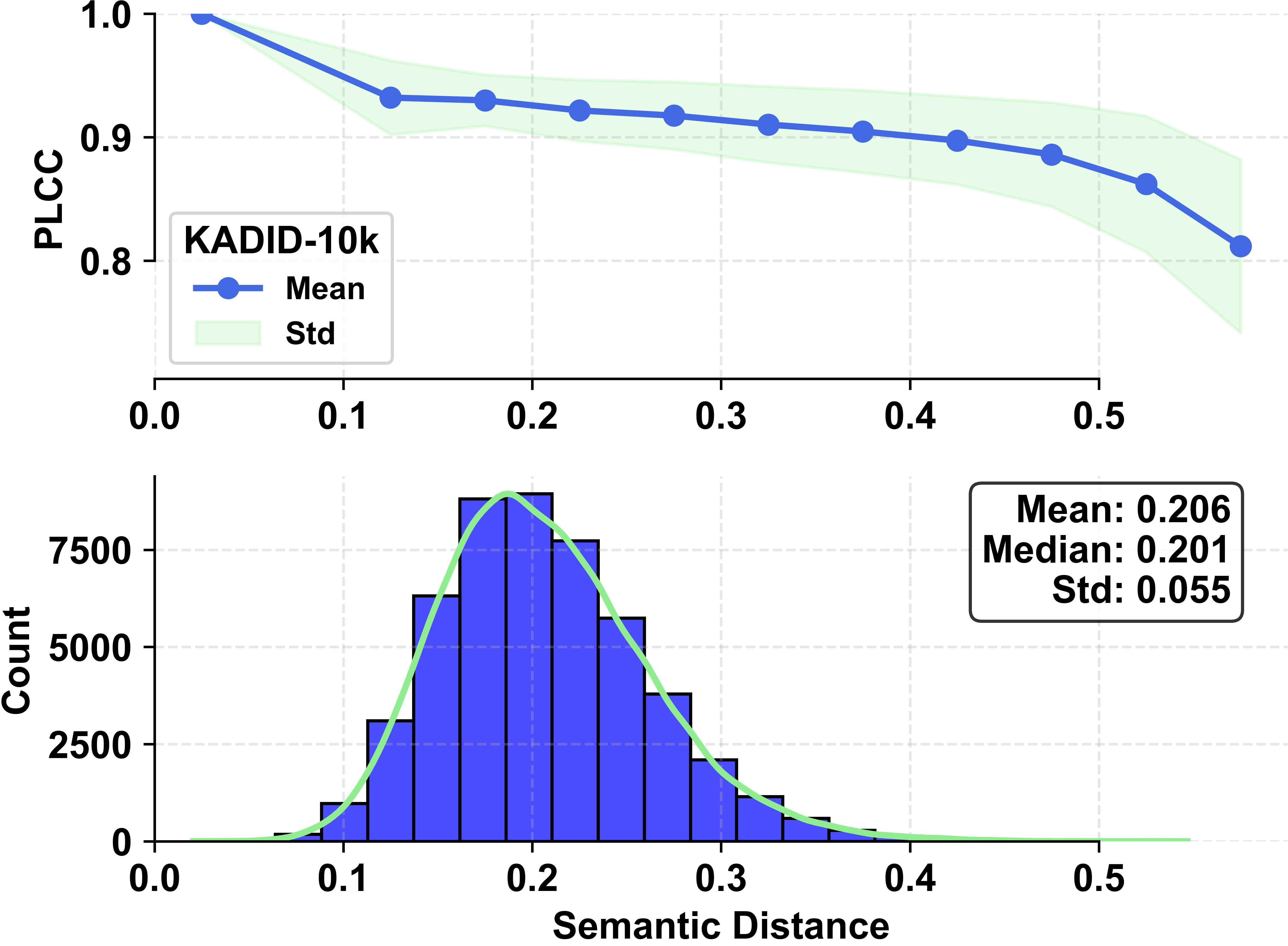}  
	\caption{Semantic distance analysis and dataset validation. Upper: PLCC distribution across semantic distances for KADID-10k (mean and standard deviation). Lower: Distribution of semantic distances in the proposed SAQT-IQA dataset.}  %\vspace{-1mm} %($\mu$=0.206, $\sigma$=0.055)
	\label{fig:plcc_mean_std} 
\end{figure}

\begin{figure*}[!tp]
	\centering
	\includegraphics[scale=.55]{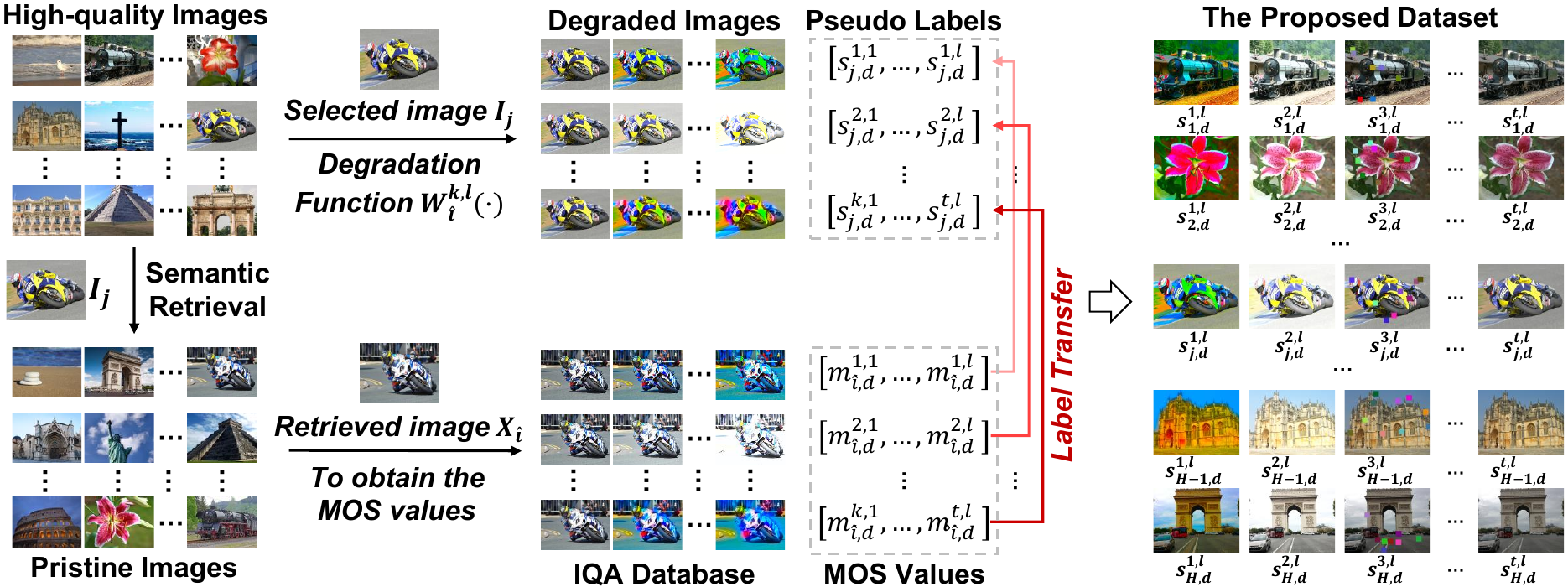} 
	\caption{The procedure of dataset construction based on the proposed semantic-aligned quality transfer method.}  
	\label{fig:dataset} 
\end{figure*}
\subsubsection{Semantic-Aligned Quality Transfer~(SAQT)-IQA dataset} The process of constructing the dataset is illustrated in Fig.~\ref{fig:dataset}, which involves the following three steps: Each high-quality image $Y_j$ and candidate pristine image $X_i$ are mapped into a high-dimensional semantic space through a feature extraction function $\mathcal{F}$, which is based on a pre-trained ResNet101~\cite{resnet}, excluding the final linear layer. The extracted feature vectors $\mathcal{F}(X_i)$ and $\mathcal{F}(Y_j)$ belong to the space $\mathbb{R}^{2048}$. These vectors capture high-level semantic information for the measurement of content similarity between images. \textbf{(b)} Content similarity determination: To quantify the semantic similarity between $Y_j$ and the set of candidate pristine images $\{X_i\}_{i=1}^P$, we define a semantic distance function $d_s(\cdot)$:
\begin{equation}
	d_s(\mathcal{F}(X_i), \mathcal{F}(Y_j)) = 1 - \frac{\langle\mathcal{F}(X_i) \cdot \mathcal{F}(Y_j)\rangle}{||\mathcal{F}(X_i)||_2 \cdot ||\mathcal{F}(Y_j)||_2},
\end{equation}
where $\langle\cdot,\cdot\rangle$ denotes the inner product and $||\cdot||_2$ is the Euclidean norm. The image $X_i$ that minimizes the semantic distance to $Y_j$ is selected as the most semantically similar pristine image:
\begin{equation}
	X_{\hat{i}} = \arg\min\limits_{\footnotesize X_i \in \{X_i\}_{i=1}^P} d_s(\mathcal{F}(Y_j), \mathcal{F}(X_i)),
\end{equation}
This step identifies the pristine image $X_{\hat{i}}$ that most closely matches the content of $Y_j$ in terms of semantic features. \textbf{(c)} Quality transfer operator: Once the most semantically similar pristine image $X_{\hat{i}}$ is identified, the next step involves transferring the quality labels from the pristine image to the high-quality image $Y_j$. This is formalized using the quality transfer function $\mathcal{T}$, which maps the high-quality image $Y_j$ to its corresponding degraded version $y^{t,l}_{j,d}$ and assigns the appropriate quality label $s^{t,l}_{j,d}$:

\begin{equation}
	\mathcal{T}: (Y_j, X_{\hat{i}}) \mapsto \left\{
		\begin{aligned}
			& \text{Degradation:} \quad y^{t,l}_{j,d} = \mathcal{D}_{t,l}^{\hat{i}}(Y_j) \\
			& \text{Quality Label:} \quad s^{t,l}_{j,d} = m^{t,l}_{\hat{i},d}
		\end{aligned}
		\right.
\end{equation}
Here, $\mathcal{D}_{t,l}^{\hat{i}}(Y_j)$ denotes the degradation operation applied to $Y_j$, producing a distorted version $y^{t,l}_{j,d}$. The MOS value $m^{t,l}_{\hat{i},d}$, associated with the degraded version $x^{t,l}_{\hat{i},d}$ of $X_{\hat{i}}$, is then transferred to annotate the degraded image $y^{t,l}_{j,d}$. Finally, the generated dataset can be represented as $\mathcal{Y}=\left\{(\textbf{Y}_{j,d},\textbf{S}_{j,d})\right\}_{j=1}^H$, where $\textbf{Y}_{j,d}=\left\{y^{t,l}_{j,d}|t\in  \mathcal{T}, l \in  \mathcal{L}\right\}$ and $\textbf{S}_{j,d}=\left\{s^{t,l}_{j,d}|t\in  \mathcal{T}, l \in  \mathcal{L}\right\}$.

The effectiveness of the SAQT-IQA dataset is further validated through an analysis of semantic distances and their correlation with MOS values. Specifically, the relationship between semantic distances and PLCC values is examined by partitioning the distance range into 0.07 intervals and calculating the mean and standard deviation within each interval based on the results shown in Fig.~\ref{fig:analysis_c}. In Fig. ~\ref{fig:plcc_mean_std}, the upper plot illustrates this correlation of KADID-10k, where a gradual decline in correlation is observed as semantic distance increases. Importantly, the majority of PLCC values remain elevated ($>$0.9) for semantic distances below ($<$0.4). The lower plot presents the semantic distance distribution of our constructed dataset, exhibiting an normal distribution ($\mu$=0.206, $\sigma$=0.055). This distribution serves two purposes: image pairs with distances below the mean $\mu$ promote label accuracy due to semantic proximity, while those above the mean $\mu$ enhance content diversity. Crucially, PLCC values stay consistently high ($>$0.9) across these distances, confirming the reliability of our label transfer strategy and validating the effectiveness of the constructed dataset. The proposed SAQT method generates quality scores that closely align with subjective human annotations, considering the influence of image content on quality perception. In contrast, the FR-IQA approach largely neglects high-level semantics and disregards content-related perceptual factors in quality measurement.

 \begin{figure*}[tp!]
	\centering
	\includegraphics[scale=.72]{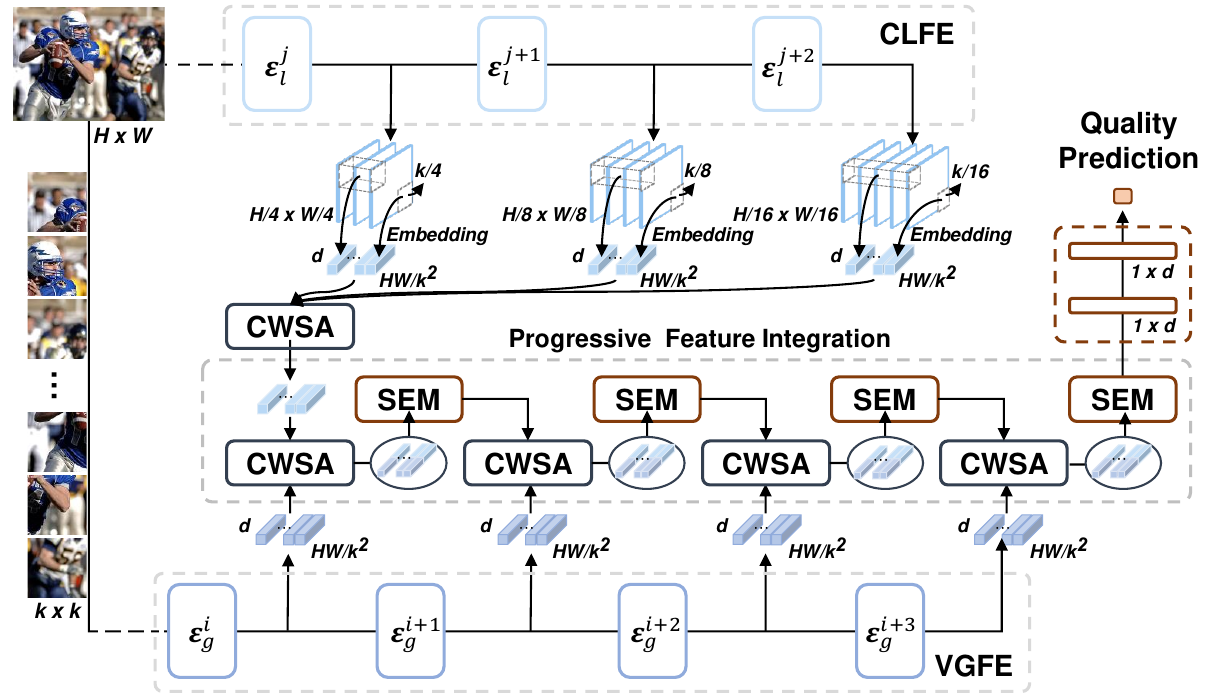} \vspace{-2mm}
		\caption{The framework of the proposed GlintIQA. The input image is processed by VGFE and CLFE, followed by feature alignment and progressive integration using interactively stacked CWSA and SIEM. The resulting multi-grained representation is then fed to an MLP for quality prediction.}
	\label{fig:framework}
	%\vspace{-1mm}
\end{figure*}

\section{The Proposed GlintIQA}\label{methodology}

As shown in Fig.\ref{fig:framework}, the proposed GlintIQA operates in three stages: global and local feature extraction (section \ref{Feature_Extraction}) via hybrid structures, \textit{i.e.}, ViT-based global feature extractor (VGFE) and CNNs-based local feature extractor (CLFE); global and local feature integration (section \ref{Feature_Integration}) using channel-wise self-attention (CWSA) and spatial interaction enhancement module (SIEM) for multi-grained representations; and image quality prediction (section \ref{Prediction}) with MLP to predict the perceptual quality score.

\subsection{Global and Local Feature Extraction} \label{Feature_Extraction}

\subsubsection{VGFE} 
%\noindent\textbf{1)~Global feature extractor (GFE):} 
Global features are indispensable for image quality evaluation, providing holistic contextual information. While convolutional operations focus on local features, the VGFE utilizes the power of self-attention mechanisms to capture long-range relationships between image regions. We utilize four vision Transformer blocks~\cite{vit}, where $i$-th block denote as $\mathcal{E}_g^i$ with parameters $\omega_g^j$.  Given an image $\textbf{x} \in \mathbb{R}^{3 \times H \times W}$, the image is partitioned into non-overlapping patches that precede global dependency modeling. Subsequently, a patch embedding layer maps these patches to individual tokens, typically achieved through convolution and flatten operation, which can be represented as: 

\begin{equation} \label{patch_embedding}
	\textbf{f}_{g}^{e} =   \psi_{\tiny f}\left(\mathcal{C}\left(\textbf{x};~3,d,k\right)\right)^\text{T},
\end{equation}  
where the notation $\mathcal{C}\left(\textbf{x}; 3,d,k\right)$  denotes a 2D convolutional layer that takes an input $\textbf{x}$ with 3 channels and transforms it into an output feature map with $d$ dimensions, using a kernel size of  $k\times k$ and a stride of $k$. The function ${{\Large \psi}}_{\tiny f}$ represents the flattening operation that collapses the output of the convolutional layer along the spatial dimensions, yielding $\textbf{f}_{g}^{e} \in \mathbb{R}^{d \times \frac{HW}{k^2}}$. Subsequently, $\textbf{f}_{g}^{e}$ is transposed to obtain features within $\mathbb{R}^{\frac{HW}{k^2} \times d}$, where each patch of size $k \times k$ is represented by a vector of length $d$, referred to as a token. For simplicity, we have omitted the description of the class token and positional embedding used in the original ViT architecture.

\begin{figure}[t!]
	\centering
	\subfigure[Spatial dimensionality reduction of ViT]{%
		\begin{minipage}[t]{0.9\linewidth}
			\centering
			\includegraphics[width=\linewidth]{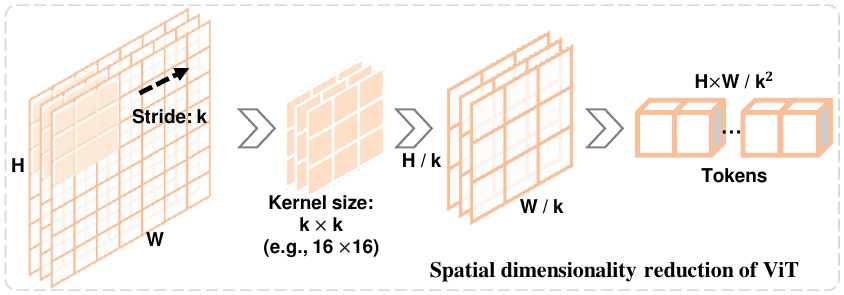}\noindent \vspace{-2mm}
			\label{fig:cnn_vit1}
		\end{minipage}%
	} %\vspace{-1mm}
	
	\subfigure[Spatial dimensionality reduction of CNNs]{%
		\begin{minipage}[t]{0.9\linewidth}
			\centering
			\includegraphics[width=\linewidth]{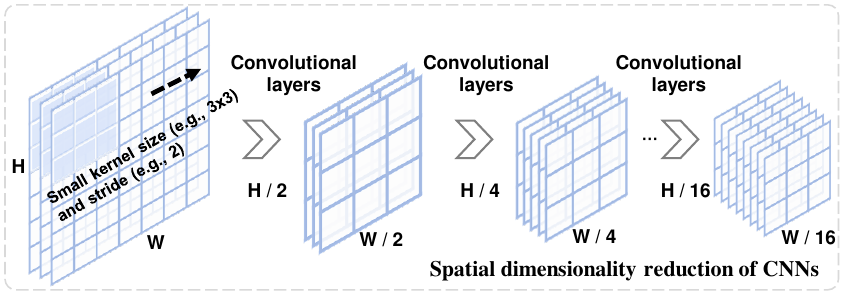}\noindent\vspace{-2mm}
			\label{fig:cnn_vit2}
		\end{minipage}
	}%\vspace{-1mm}
	\caption{Comparison of spatial dimensionality reduction in ViT and CNNs. (a): In ViT, the spatial dimensionality reduction occurs through the patch embedding process, which subdivides an input image of dimensions $H \times W$  into non-overlapping patches using a convolutional layer with a kernel size $k \times k$ and stride $k$ (\textit{e.g.}, $k=16$), resulting in tokens of size $H/k \times W/k$. (b): Spatial dimensionality reduction in CNNs involves applying successive convolutional layers with small kernel sizes (\textit{e.g.},  $3 \times 3$) and strides (\textit{e.g.}, 2), progressively halving the spatial dimension.}
	\label{cnn_vits}
\end{figure}
\begin{figure*}[tbp!]
	\centering
	\subfigure[The structure of CWSA]{%
		\begin{minipage}[t]{0.48\linewidth}
			\centering
			\includegraphics[width=\linewidth]{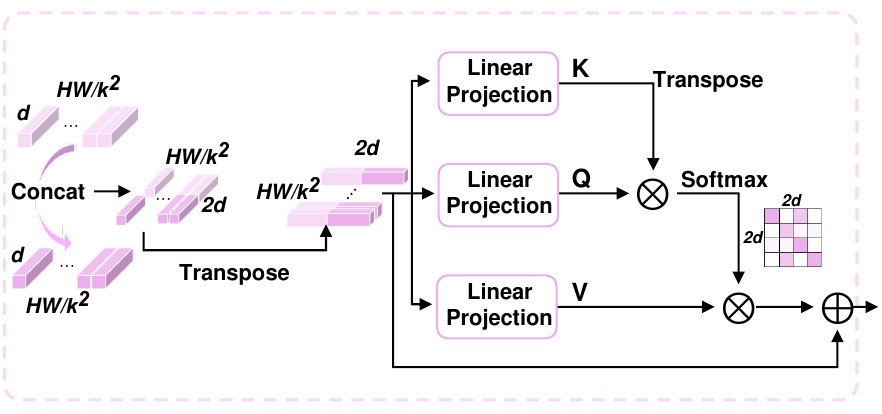}\noindent\vspace{-2mm}
			\label{fig:cwsa}
		\end{minipage}%
	}%
	\subfigure[The structure of SIEM]{%
		\begin{minipage}[t]{0.48\linewidth}
			\centering
			\includegraphics[width=\linewidth]{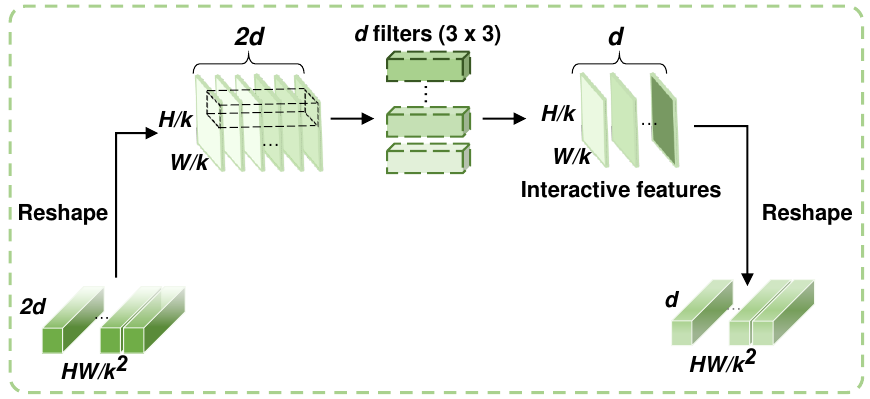}\noindent\vspace{-2mm}
			\label{fig:lem}
		\end{minipage}
	}%
	
	\caption{Structures of (a) channel-wise Self-attention (CWSA) and (b) Spatial Interaction Enhancement Module (SIEM).} \vspace{-2mm}
	\label{two_module}  
\end{figure*}

Subsequently, these tokens are input into a Transformer block for extracting global features. The output of the $i$-th block can be expressed as:
\begin{equation}
	\textbf{f}_g^i = \mathcal{E}_g^i\left(\textbf{x};~\omega_g^i\right),
\end{equation}
where the extracted features at different scales maintain consistent dimensionality, residing in $\mathbb{R}^{\frac{HW}{k^2} \times d}$.

Although the patch embedding process, as shown in Eq.~\ref{patch_embedding}, enables subsequent models to capture global context, it may result in the loss of fine-grained details within patches. As shown in Fig.~\ref{fig:cnn_vit1}, patch embedding discretizes the input image into a sequence of visual tokens, significantly reducing spatial dimensions in a single step at the model input. This discretization inevitably leads to a loss of fine-grained details and local spatial information, as the individual tokens correspond to low-resolution patches rather than individual pixels. To address this limitation, we exploit a local feature extractor consisting of a series of convolutional layers, as depicted in Fig.~\ref{fig:cnn_vit2}. This extractor uses small convolutional kernels and strides to progressively reduce the dimensions of the image,  which can preserve certain local image details in the extracted features. Moreover, the integrated CNNs-based structure can mitigate the inherent lack of inductive bias in ViTs. By leveraging the strengths of both CNNs and ViTs, the hybrid model effectively captures both global and local features, ensuring a more accurate and robust representation of image quality.

\subsubsection{CLFE}
Local features~(\textit{e.g.}, edges and textures) reside within small image regions and provide valuable insights into the nature and severity of local distortions. To extract these features, the CLFE utilizes three residual blocks from ResNet50~\cite{resnet}, denoted as $\mathcal{E}_l^j$ with parameters $\omega_l^j$, where $j$ represents the index of the blocks. Each block yields features at distinct scales, downsampled by factors of $2^{j+1}$ relative to the original image.

To ensure the local and global features are aligned in size before further integration, a multi-kernel embedding scheme is employed that adjusts the kernel size and stride based on the downsampling ratio of the input features. Specifically, the processing of the $j$-th output feature can be represented as:
\begin{equation}
	\textbf{f}_l^j ={\Large \psi}_{\tiny f} \left( \mathcal{C}\left({\mathcal{E}}_l^j\left(\textbf{x};  \omega_l^j\right);~c_j, d, k/2^{j+1}\right)\right)^\text{T},
\end{equation}
where $c_j$ denotes the feature channels of the output from the $j$-th residual block. To maintain spatial alignment of features at the $j$-th level, which has undergone a $2^{j+1}$-fold reduction in size, convolutional kernels and strides are adaptively set to $k/{2^{j+1}}$, ensuring precise correspondence with the original image region dimensions of $k \times k$.

The outputs of the three side blocks are concatenated along the channel dimension. Then, we model the channel weights for multi-scale features through channel-wise self-attention. Following this, a linear layer is utilized to map these features to a dimensionality of $d$. These processes can be represented as:
\begin{equation}
	\textbf{f}_l = \mathcal{P}\left( \mathcal{M}_\text{CWSA}\left(\textbf{f}_l^{j-1} \oplus \textbf{f}_l^j \oplus \textbf{f}_l^{j+1}\right);~3d, d\right) ,
\end{equation}
where $\textbf{f}_l \in \mathbb{R}^{\frac{HW}{K^2} \times d}$ and $\oplus$ denotes a concatenation operation, and $\mathcal{P}(\cdot;3d,d)$ denotes a linear projection layer with input and output dimensions of $3d$ and $d$, respectively. The CWSA module, denoted as  $\mathcal{M}_\text{CWSA}$, is described in detail in the following subsection.

\subsection{Global and Local Feature Integration}\label{Feature_Integration}
Integrating local and global features presents several challenges. Local features from CNNs and global features from ViTs operate at different scales, capturing fine-grained and coarse-grained details, respectively. A naive fusion of these multi-scale features risks diluting their distinct advantages and fails to capture the intricate dependencies between them. Additionally, local and global features contribute unequally across channels, with some local channels capturing critical texture details and others emphasizing high-level structural information. Thus, a more effective fusion mechanism is needed to integrate these multi-scale features while preserving their complementary nature. A dynamic, adaptive weighting mechanism is crucial to prioritize the most informative channels, ensuring a balanced and efficient fusion of local and global representations.

\subsubsection{Progressive Feature Integration}
Direct fusion of multi-granular features risks losing their complementary strengths by merging fine-grained and coarse-grained information simultaneously. To address this, a progressive integration strategy is employed, gradually merging features across layers. This approach allows the model to refine both local and global representations, optimizing them step-by-step across channels and spatial dimensions for more effective feature integration. Specifically, we start by concatenating the fine-grained feature $\textbf{f}_l$ with the coarse-grained global feature at the initial scale. Then, a stacked combination of CWSA and a SIEM is used to progressively fuse features in a hierarchical manner. The feature integration process can be concisely represented as: 
\begin{equation} \label{eq_integration}
	\left\{
	\begin{aligned}
		\textbf{z}_{i} &= \left( \textbf{f}_i \oplus \textbf{f}_g^i\right)^{\text{T}}; \\
		\hat{\textbf{z}}_{i} &=\mathcal{M}_{\text{CWSA}}\left(\textbf{z}_{i} \right); \\
		\textbf{f}_{i+1} &= \mathcal{M}_{\text{SIEM}}\left( \hat{\textbf{z}}_{i}\right).
	\end{aligned}
	\right.
\end{equation}

For $i=1$, $\mathbf{f}_i$ represents the output feature of the CLFE, \textit{i.e.}, $\mathbf{f}_l$. For other values of $i$, $\mathbf{f}_i$ is the output feature of the SIEM, denoted as $\mathcal{M}_\text{SIEM}$. 

\subsubsection{CWSA}~Given the varying contributions of different channels in local and global features, CWSA is designed to dynamically learn and adjust the relative importance of each channel. This ensures that channels containing critical local texture details or key global structural information are appropriately emphasized. Using CWSA, the model computes the relevance of each channel to the task, reinforcing informative channels while down-weighting redundant ones. This adaptive weighting mechanism addresses the challenge of channel-wise importance disparity, enabling a more effective and balanced fusion process. 
%As illustrated in Fig.~\ref{fig:cwsa}, CWSA concatenates two inputs to obtain the feature $\textbf{z}_{i}$ as shown in Eq.~\ref{eq_integration}. Subsequently, the feature is transposed and subjected to a self-attention computation. Specifically, $\textbf{z}_{i}$ is projected into three representations for query, key, and value through three distinct linear projection layers, which is expressed as:
%\begin{equation}
%		\textbf{Q} = \mathcal{P}^Q(\textbf{z}_{i};d,d);\textbf{K}=\mathcal{P}^K(\textbf{z}_{i};d,d); \textbf{V} = \mathcal{P}^V(\textbf{z}_{i};d,d).
%\end{equation}
% 
%The softmax scores quantify the channel-wise significance, which get applied to the scaled dot product of queries ($\textbf{Q}$) and keys ($\textbf{K}$), normalized by $\sqrt{d}$ ($d$ being the dimension). The CWSA output is the weighted summation of values ($\textbf{V}$) based on these attention scores, combined with the previous input feature $\textbf{z}_{i}$, which is expressed as: 
%\begin{equation}
%		\hat{\textbf{z}}_{i} = \text{Softmax}(\text{\textbf{Q}} \cdot \text{\textbf{K}}^\text{T} / \sqrt{d}) \cdot \text{\textbf{V}} + \textbf{z}_{i},
%\end{equation}
%where $\text{Softmax}(\cdot)$ represents the softmax function applied to the input, which normalizes the values into a probability distribution over the output dimensions.
As illustrated in Fig.\ref{fig:cwsa}, CWSA concatenates two inputs to obtain the feature $\textbf{z}_{i}$ as shown in Eq.\ref{eq_integration}. Subsequently, the feature is transposed and subjected to a self-attention computation through the following equations:
\begin{align}
	\textbf{Q} &= \mathcal{P}^Q(\textbf{z}_i;d,d),
	\textbf{K} = \mathcal{P}^K(\textbf{z}_i;d,d),
	\textbf{V} = \mathcal{P}^V(\textbf{z}_i;d,d), \\
	\hat{\textbf{z}}_{i} &= \text{Softmax}(\text{\textbf{Q}} \cdot \text{\textbf{K}}^\text{T} / \sqrt{d}) \cdot \text{\textbf{V}} + \textbf{z}_{i},
\end{align}
where $\textbf{z}_{i}$ is first projected into three distinct representations (query $\textbf{Q}$, key $\textbf{K}$, and value $\textbf{V}$) through separate linear projection layers $\mathcal{P}^Q$, $\mathcal{P}^K$, and $\mathcal{P}^V$. The channel-wise significance is then quantified through softmax scores applied to the scaled dot product of queries and keys, normalized by $\sqrt{d}$ ($d$ being the dimension). The final CWSA output $\hat{\textbf{z}}_{i}$ is obtained by combining the attention-weighted values with the residual connection of the input feature $\textbf{z}_{i}$, where $\text{Softmax}(\cdot)$ normalizes the values into a probability distribution over the output dimensions.

\subsubsection{SIEM}~SIEM follows channel-wise attention to enhance spatial relationships between local and global features. By reshaping and applying convolutional operations, SIEM fosters deeper interactions across multiple scales, effectively capturing spatial dependencies and complementing CWSA to produce a representation that is both channel-optimized and spatially robust. The module operates on the input tensor $\hat{\textbf{z}}_i$, as illustrated in Fig.\ref{fig:lem}. The computational process is expressed as:
\begin{equation}
	\textbf{f}_{i+1} = \psi_f \left(\mathcal{C}\left( \varphi_r\left( \hat{\textbf{z}}_{i}\right);~2d, d, 3\right)\right),
\end{equation}
where the notation ${\Large \varphi}_{\tiny r}$ indicates the operation of reshaping the token vectors into 2D feature maps. The convolutional layer $\mathcal{C}(\cdot;2d, d, 3)$ employs a $3\times 3$ filter with a stride of 1. Finally, the output of the convolutional layer is reshaped back to a flattened tensor, aligning it with the original input dimensions.
\begin{table}[t]
	\centering
	\caption{The Summary of IQA Databases for NR-IQA Models.} 
	\fontsize{25pt}{25pt}\selectfont%
	\renewcommand{\arraystretch}{1.}
	\resizebox{3.5 in}{!} 	{
		\begin{tabular}{cccccc}
			\toprule
			\textbf{Database}&	\textbf{Database} & \textbf{Ref. Image} & \textbf{Dist. Type} & \textbf{Dist. Image} &\textbf{Image}\tabularnewline
			\textbf{Type} &	\textbf{Name} & \textbf{Count} & \textbf{Count} & \textbf{Count} & \textbf{Resolution}\tabularnewline
			\midrule
			\multirow{5}{*}{\rotatebox{90}{Synthetic}}&	LIVE \cite{live} 	 & 29 & 5 & 779 &480$\times$720 $\sim$ 768$\times$512 \tabularnewline
			&	CSIQ \cite{csiq} 	 & 30 & 6 & 866 &512$\times$512 \tabularnewline
			&		TID2013 \cite{tid2013} & 25 & 24 & 3,000 &384$\times$512 \tabularnewline
			&		KADID-10k \cite{kadid-10k} & 81 & 25 & 10,125 &384$\times$512\tabularnewline
			&		LIVE-MD \cite{livemd}& 15 & 2 & 405 &720$\times$1280 \tabularnewline
			%PIPAL\cite{livemd}& 250 & 40 & 29,000 &288$\times$288& MOS&synthetic\tabularnewline
			\midrule
			\multirow{5}{*}{\rotatebox{90}{Authentic}} &		BID~\cite{bid}&-&-&585&960$\times$1280 $\sim$ 1704$\times$2272 \tabularnewline
			&		LIVEC~\cite{livec}&-&-&1,162&500$\times$500 $\sim$ 640$\times$960 \tabularnewline
			&		KonIQ-10k~\cite{koniq-10k}&-&-&10,073&384$\times$512/768$\times$1024 \tabularnewline
			&		SPAQ~\cite{spaq}&-&-&11,125&1080$\times$1440$\sim$6556$\times$3744 \tabularnewline
			&		LIVEFB~\cite{livefb}&-&-&39,810&160$\times$186 $\sim$ 660$\times$1200\tabularnewline
			\bottomrule
		\end{tabular}	\vspace{-1mm}
	} 
	\label{datasets_summary}
\end{table}

\subsection{Image Quality Prediction}\label{Prediction}
The quality score prediction is performed via an MLP architecture with two linear projection layers. The feature map $\textbf{f}$ from the SIEM module is spatially averaged to obtain a $d$-dimensional vector $\hat{\textbf{f}} \in \mathbb{R}^{1 \times d}$, forming a compact quality representation. This vector is then processed through two linear projections with GELU activation and dropout regularization to generate the final quality score $s$:
 
\begin{align} 
  s =  \mathcal{P}\left( \mathcal{P}(\hat{\textbf{f}};~d, d); d,1\right).
\end{align}
where $\mathcal{P}$ denotes a linear projection layer. For simplicity, the specific details of the activation function and dropout layer are omitted from the above equation. Ultimately, $l_1$ loss function is utilized to enable end-to-end model parameter updates.

\section{Experiment} \label{experiment}

This section begins by outlining the experimental setup (section \ref{subsec:setting}). Subsequently, we compare the experimental results against state-of-the-art NR-IQA methods across ten IQA databases, encompassing both single-dataset (section \ref{Comparisons_individual}) and cross-dataset (section \ref{Comparisons_cross}) evaluations. Finally, we perform comprehensive ablation studies to further assess the effectiveness of our approach (section \ref{Ablation}).

\begin{figure}[tbp!]
	\centering
	\includegraphics[scale=.44]{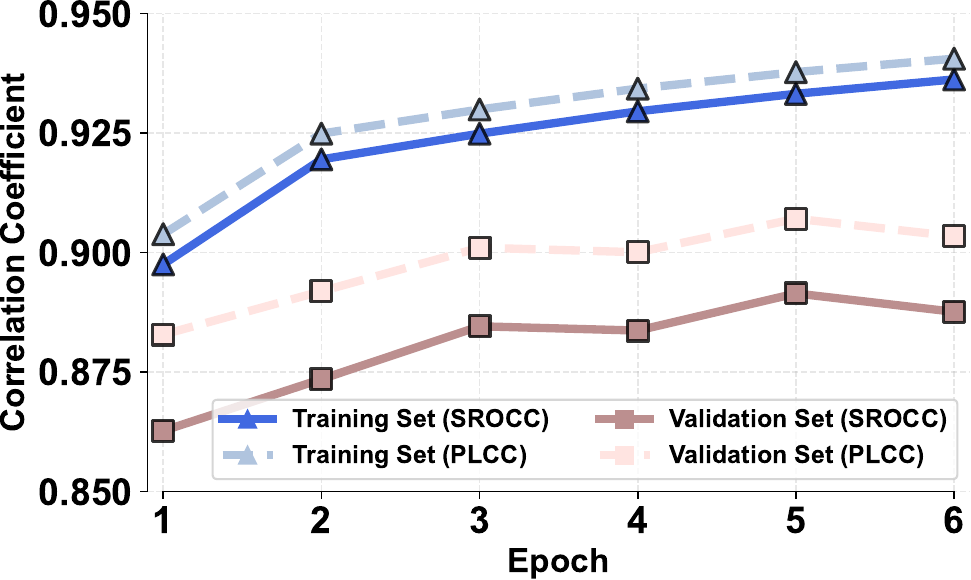} 
	\caption{The pre-training results of the proposed SAQT-IQA dataset.}  \vspace{-1mm}
	\label{pretrain} 
\end{figure}
\subsection{Experimental Settings} \label{subsec:setting}
\subsubsection{Databases and protocols}
A comprehensive evaluation of the proposed model was conducted across ten IQA benchmark databases, including five synthetic distortion datasets, specifically LIVE~\cite{live}, CSIQ~\cite{csiq}, TID2013~\cite{tid2013}, KADID-10k~\cite{kadid-10k}, and LIVE-MD~\cite{livemd}, alongside five authentic distortion datasets, namely BID~\cite{bid}, CLIVE~\cite{livec}, KonIQ-10k~\cite{koniq-10k}, SPAQ~\cite{spaq}, and LIVEFB~\cite{livefb}. Within the synthetic distortion datasets, the LIVE-MD dataset features images with multiple distortion types,  whereas the remaining databases primarily focus on images with a single distortion type. Synthetic distortion datasets are constructed by algorithmically degrading original images, resulting in minimized content variation but a wide range of artificial distortions. In contrast, authentic distortion datasets comprise images captured in real-world settings, exhibiting substantial content diversity from varied scenes, objects and conditions encountered. The details of sample size, image resolution, and label types for each database are summarized in Table~\ref{datasets_summary}. The evaluation of NR-IQA methods employs two widely-adopted correlation metrics: Spearman's rank-order correlation coefficient (SROCC) and Pearson's linear correlation coefficient (PLCC). SROCC quantifies the monotonic association between predicted scores and ground truth, while PLCC measures their linear relationship. %Specifically, SROCC computes the squared disparity between the ranks of predicted quality scores and MOS values. On the other hand, PLCC quantifies linear relationship between predicted quality scores and MOS values by calculating the covariance divided by the product of their standard deviations.

\begin{table*}[htp!]
	\centering
	\caption{Benchmarking performance against existing NR-IQA models on authentic distortion databases, where the best and second-best performances are indicated by \textbf{black} bolded and \textbf{\textcolor{deepgray}{gray}} bolded entries, respectively.} 
	\fontsize{14.pt}{14.pt}\selectfont%
	\renewcommand{\arraystretch}{1.}
	\resizebox{6.8 in}{!}{
		\begin{tabular}{ccccccccccccc}
			\toprule
			\multirow{2}{*}{\textbf{Method}} & \multicolumn{2}{c}{\textbf{CLIVE}\cite{livec}} & \multicolumn{2}{c}{\textbf{KonIQ-10k}\cite{koniq-10k}} & \multicolumn{2}{c}{\textbf{BID}\cite{bid}}& \multicolumn{2}{c}{\textbf{SPAQ}\cite{spaq}}  & \multicolumn{2}{c}{\textbf{LIVEFB}\cite{livefb}}&\multicolumn{2}{c}{\textbf{Average}}\\
			\cmidrule(r){2-3}\cmidrule(r){4-5}\cmidrule(r){6-7}\cmidrule(r){8-9} \cmidrule(r){10-11} \cmidrule(r){12-13}
			\multirow{2}{*}{} & \textbf{SROCC} & \textbf{PLCC} & \textbf{SROCC} & \textbf{PLCC} & \textbf{SROCC} & \textbf{PLCC} & \textbf{SROCC} & \textbf{PLCC} & \textbf{SROCC} & \textbf{PLCC}&\textbf{SROCC} & \textbf{PLCC} \tabularnewline
			\midrule
			WaDIQaM-NR\cite{WaDIQaM} &0.692 &0.730 &0.729& 0.754& 0.653 &0.636 &0.840 &0.845 &0.435& 0.430&0.670&0.679\tabularnewline
			SFA~\cite{sfa}&0.812& 0.833& 0.856& 0.872& 0.826& 0.840&0.906& 0.907& 0.542& 0.626&0.788
			&0.816\tabularnewline
			MUSIQ~\cite{musiq} &0.702& 0.746 & 0.916&0.928&-& -& \textbf{\textcolor{deepgray}{\underline{0.918}}}& 0.921&-&-&-&-\tabularnewline 
			DB-CNN\cite{dbcnn} &0.844 &0.862 &0.878& 0.887 &0.845&0.859&0.910 &0.913& 0.554& 0.652&0.806 
			&0.835\tabularnewline
			HyperIQA\cite{hyperiqa} &0.859&0.882& 0.906& 0.917& 0.869 &0.878 &0.911 & 0.915 & 0.544&0.602&0.818
			&0.839\tabularnewline
			
			TReS\cite{tres}&0.846& 0.877 &  0.915&0.928 &0.855& 0.871&0.917 &0.913& 0.554& 0.625&0.817 
			&0.843\tabularnewline
			REQA\cite{reqa} &0.865& 0.880& 0.904& 0.916&\textbf{\textcolor{deepgray}{\underline{0.874}}} & 0.886 &-& -& -& -&-&-\tabularnewline
			TempQT\cite{temqt}  & 0.870 & 0.886 & 0.903 & 0.920 &0.868&0.902 &0.916&0.922 & 0.561 &0.646&0.824&0.855 \tabularnewline
			DEIQT~\cite{deiqt}	&\textbf{\textcolor{deepgray}{\underline{0.875}}} &\textbf{\textcolor{deepgray}{\underline{0.894}}}&\textbf{\textcolor{deepgray}{\underline{ 0.921}}}& \textbf{\textcolor{deepgray}{\underline{0.934}}} &\textbf{0.889}&\textbf{\textcolor{deepgray}{\underline{0.903}}}&\textbf{0.919} &\textbf{\textcolor{deepgray}{\underline{0.923}}}&  \textbf{0.571} &\textbf{0.663} &\textbf{\textcolor{deepgray}{\underline{0.835}}}&\textbf{\textcolor{deepgray}{\underline{0.863}}}\tabularnewline
			\midrule
			
			\rowcolor{Tan!10}GlintIQA& \textbf{0.891}& \textbf{0.907}&\textbf{0.936}&\textbf{0.946}&\textbf{0.889}&\textbf{0.906}&\textbf{0.919}&\textbf{0.924}&\textbf{\textcolor{deepgray}{\underline{0.566}}} & \textbf{\textcolor{deepgray}{\underline{0.653}}}&\textbf{0.840}&\textbf{0.867} \tabularnewline
			\bottomrule
		\end{tabular}
	}
	\label{authentic_dataset} %\vspace{-1.em}
\end{table*}

\begin{table*}[htp!]
	\centering
	\caption{Comparison to state-of-the-art NR-IQA models on five synthetic IQA databases, where the best and second-best performances are indicated by \textbf{black} bolded and \textbf{\textcolor{deepgray}{gray}} bolded entries, respectively. GlintIQA* was fine-tuned on the respective dataset using the pre-trained model on the proposed SAQT-IQA dataset.} 
	\fontsize{14.pt}{14.pt}\selectfont%
	\renewcommand{\arraystretch}{1.}
	\resizebox{6.8 in}{!}{
		\begin{tabular}{ccccccccccccc}
			\toprule
			\multirow{2}{*}{\textbf{Method}} & \multicolumn{2}{c}{\textbf{LIVE}\cite{live}} & \multicolumn{2}{c}{\textbf{CSIQ}\cite{csiq}} & \multicolumn{2}{c}{\textbf{TID2013}\cite{tid2013}}& \multicolumn{2}{c}{\textbf{KADID-10k}\cite{kadid-10k}}  & \multicolumn{2}{c}{\textbf{LIVEMD}\cite{livemd}}&\multicolumn{2}{c}{\textbf{Average {(  w/o KADID-10k)}}}\\
			\cmidrule(r){2-3}\cmidrule(r){4-5}\cmidrule(r){6-7}\cmidrule(r){8-9} \cmidrule(r){10-11} \cmidrule(r){12-13}
			\multirow{2}{*}{} & \textbf{SROCC} & \textbf{PLCC} & \textbf{SROCC} & \textbf{PLCC} & \textbf{SROCC} & \textbf{PLCC} & \textbf{SROCC} & \textbf{PLCC} & \textbf{SROCC} & \textbf{PLCC}&\textbf{SROCC} & \textbf{PLCC} \tabularnewline
			\midrule
			MEON~\cite{meon}&0.951&0.955&0.852&0.864&0.808&0.824&0.604&0.691&0.924&0.940&0.884&0.896
			\tabularnewline
			SGDNet\cite{sgdnet} & 0.969 & 0.965 & 0.883 & 0.903 & 0.843 & 0.861 & - & - &-&-&-&-\tabularnewline
			CaHDC\cite{cahdc} & 0.965 & 0.964 & 0.903 & 0.914 & 0.862 & 0.878 & - & - & 0.927 & \textbf{0.950} &-&-\tabularnewline
			HyperIQA\cite{hyperiqa}  &0.962&0.966&0.923& 0.942&0.840& 0.858&0.852  &0.845 &\textbf{\textcolor{deepgray}{\underline{0.939}}}&0.938&0.916&0.926
			\tabularnewline
			DB-CNN\cite{dbcnn} &0.968&0.971 &0.946& 0.959&0.816 &0.865 &0.851 &0.856&0.927&0.934 &0.914&0.932
			\tabularnewline
			TReS\cite{tres} &0.969&0.968& 0.922& 0.942&0.863&0.883&0.859 &0.858&0.930&0.932&0.921&0.931
			\tabularnewline
			AIGQA\cite{aigqa} & 0.960 & 0.957 & 0.927 & 0.952 & 0.871 &0.893 &0.864&0.863& 0.933 & \textbf{\textcolor{deepgray}{\underline{0.947}}}&0.923&0.937 \tabularnewline
			
			TempQT\cite{temqt}  & \textbf{\textcolor{deepgray}{\underline{ 0.976 }}}& 0.977 &  \textbf{\textcolor{deepgray}{\underline{0.950}}} & 0.960 & 0.883 & 0.906 &- &- & 0.933 &0.939&\textbf{\textcolor{deepgray}{\underline{0.936}}}&\textbf{\textcolor{deepgray}{\underline{0.946 }}}\tabularnewline
			DEIQT~\cite{deiqt}&	\textbf{0.980}	&\textbf{0.982} &0.946 &\textbf{\textcolor{deepgray}{\underline{0.963}}}&  \textbf{\textcolor{deepgray}{\underline{0.892}}}& \textbf{\textcolor{deepgray}{\underline{0.908}}}& \textbf{\textcolor{deepgray}{\underline{ 0.889}}}& \textbf{\textcolor{deepgray}{\underline{0.887}}} & 0.907 &0.930&0.931&\textbf{\textcolor{deepgray}{\underline{0.946}}} \tabularnewline
			\midrule
			\rowcolor{Tan!10}GlintIQA&  \textbf{\textcolor{deepgray}{\underline{0.976}}}& \textbf{\textcolor{deepgray}{\underline{0.980}}}&0.941&0.947&0.857&0.883&\textbf{0.906}&\textbf{0.903}&\textbf{0.942} &  0.940 &0.929&0.938\tabularnewline
			\rowcolor{Tan!10}GlintIQA* & \textbf{\textcolor{deepgray}{\underline{0.976}}}&0.978&\textbf{0.970}&\textbf{0.974}&\textbf{0.917}&\textbf{0.929}&\textbf{-}&\textbf{-}&0.935 & 0.945&\textbf{0.950}&\textbf{0.957} \tabularnewline
			\bottomrule
		\end{tabular}
	}
	\label{sythetic_dataset}
\end{table*}

\subsubsection{Implementation details} The proposed model was implemented using the PyTorch framework and evaluated on IQA databases utilizing NVIDIA RTX 3090 GPUs. To extract local fine-grained features, we employed the first three residual blocks of ResNet50, as the early stages effectively capture local details and intricate patterns of the input image, where these blocks have 256, 512, and 1024 output channels respectively. For global feature extraction, we used four transformer blocks from ViT-S/16~\cite{vit}, specifically blocks 6, 7, 8, and 9, which output 384-dimensional feature representations. The parameter $d$  was set to 384. The CLFE and VGFE modules were initialized using pre-trained weights from ImageNet, with the initial layers frozen to mitigate rapid overfitting. The Adam optimizer with a weight decay of 1e-5 was used for training. The training process spanned 300 epochs, using a batch size of 32 and an initial learning rate of 1e-5, which was controlled by a cosine annealing schedule. In training phase, data augmentation involved randomly cropping 224$\times$ 224 pixel patches from images as input to the model. Model performance was periodically evaluated every 5 epochs. For inference,  25 randomly extracted patches of  $224 \times 224$ pixels from each test image were used to compute the final quality score by averaging their predicted scores.  

Pre-training on our proposed large-scale dataset was performed using an NVIDIA A800 GPU and the Adam optimizer with 1e-5 weight decay. Due to the vast size of the dataset, the training process consists of 6 epoch, with batch size and initial learning rate set to 192 and 5e-5, respectively. The learning rate was decayed via cosine annealing scheduler for convergence. Each image was randomly cropped to 224$\times$224 pixels during pre-training. For validation, 1,000 images were randomly sampled from a pool of 50,000, with their corresponding degraded versions forming the validation set. To evaluate the test images, we randomly selected 10 patches and averaged their scores to yield a final prediction. Weights demonstrating optimal validation performance were retained. Training and validation results are presented in Fig.~\ref{pretrain}.

\subsection{Comparison with NR-IQA Methods on Individual Datasets} \label{Comparisons_individual}
To evaluate the proposed method, we conducted experiments on five authentic and five synthetic distortions datasets. For synthetic distortion datasets, pristine images were randomly partitioned into two subsets at an 8:2 ratio, with respective distorted images forming training and test sets to avoid content overlap. The authentic distortion dataset was partitioned into training and test sets in an 8:2 ratio. This process was repeated ten times with different random partitions, and the median SROCC and PLCC values were reported. As detailed in Tables~\ref{authentic_dataset} and \ref{sythetic_dataset}, we benchmarked our method against nine existing deep learning-based NR-IQA approaches, on both authentic and synthetic datasets. The results of the deep learning-based models were evaluated using the reported results in the original papers or replicated using their publicly released code.

\subsubsection{Authentic database evaluation} 

As shown in Table~\ref{authentic_dataset}, the proposed GlintIQA demonstrates consistently superior performance across five authentic distortion datasets, achieving an average SROCC of 0.840 and PLCC of 0.867. This represents improvements of 1.94\% and 1.40\% over TempQT, and 0.60\% and 0.46\% over DEIQT, respectively. Specifically, on the CLIVE, GlintIQA achieves best results, outperforming DEIQT by 1.83\% in SROCC and 1.45\% in PLCC. Similarly, for the KonIQ-10k, the proposed method achieves the highest performance, with SROCC and PLCC gains of 1.63\% and 1.28\% over the second-best method. In the BID and SPAQ datasets, both GlintIQA and DEIQT reach best performance, with our method exhibiting a superior PLCC. Additionally, GlintIQA secures competitive results on the LIVEFB. Overall, the proposed GlintIQA exhibits superior and more stable performance compared to existing NR-IQA methods. The improvement of GlintIQA can be attributed to two key design elements: firstly, the method integrates both global coarse- and local fine-grained image features, enhancing its sensitivity to various distortions. Secondly, the inclusion of CNNs, specifically CLFE and SIEM, significantly augments the model's inductive capacity to learn representations of different distortions. This enhanced inductive capability is particularly advantageous for achieving superior performance on smaller datasets, \textit{e.g.}, CLIVE and BID.
  \begin{table}[t!]
 	\caption{Summary of the statistical performance of the proposed method and existing NR-IQA methods.}% \vspace{-1.em}
 	\label{combined_table}
 	\centering
 	\setlength\tabcolsep{3pt}
 	\renewcommand\arraystretch{1}
 	
 	\begin{tabular}{c}
 		\scalebox{1.05}[1.05]{
 			\begin{tabular}{cccccc}
 				\toprule
 				\multicolumn{6}{l}{\textbf{\textit{Significance test results for authentic distortion datasets.}}} \\  
 				\midrule
 				& CLIVE & KonIQ-10k & BID &SPAQ& LIVEFB  \\  
 				\midrule
 				DB-CNN~\cite{dbcnn}   &\cellcolor{blue!10}1  &\cellcolor{blue!10}1 & \cellcolor{blue!10}1 &\cellcolor{blue!10}1 &\cellcolor{blue!10}1  \\
 				HyperIQA ~\cite{hyperiqa}   &\cellcolor{blue!10}1  &\cellcolor{blue!10}1 & \cellcolor{blue!10}1 &\cellcolor{blue!10}1 &\cellcolor{blue!10}1  \\
 				TReS~\cite{tres}   &\cellcolor{blue!10}1  &\cellcolor{blue!10}1 & \cellcolor{blue!10}1 &\cellcolor{blue!10}1 &\cellcolor{blue!10}1  \\
 				TempQT~\cite{temqt}   &  \cellcolor{blue!10}1  &\cellcolor{blue!10}1 & \cellcolor{blue!10}1 &\cellcolor{Tan!10}- &\cellcolor{blue!10}1  \\
 				DEIQT~\cite{deiqt} &  \cellcolor{blue!10}1  &\cellcolor{blue!10}1 &\cellcolor{Tan!10} - &\cellcolor{Tan!10}- &\cellcolor{red!10}0  \\
 			\end{tabular} 
 		}	\\
 		\vspace{3.em}  
 		\scalebox{.95}[.95]{
 			\begin{tabular}{cccccc}
 				\toprule
 				\multicolumn{6}{l}{\textbf{\textit{Significance test results for synthetic distortion datasets.}}}  \\ 
 				\midrule 
 				& LIVE & CSIQ & TID2013 &KAADID-10k& LIVEMD \\  
 				\midrule
 				DB-CNN~\cite{dbcnn}  &\cellcolor{blue!10}1  &\cellcolor{blue!10}1 & \cellcolor{blue!10}1 &/ &\cellcolor{blue!10}1  \\
 				HyperIQA ~\cite{hyperiqa}  & \cellcolor{blue!10}1  &\cellcolor{blue!10}1 & \cellcolor{blue!10}1 &/ &\cellcolor{Tan!10}-  \\
 				TReS~\cite{tres}      & \cellcolor{blue!10}1  &\cellcolor{blue!10}1 & \cellcolor{blue!10}1 &/ &\cellcolor{blue!10}1  \\
 				DEIQT~\cite{deiqt}    &\cellcolor{blue!10}1  &\cellcolor{blue!10}1 &\cellcolor{blue!10}1 &/ &\cellcolor{blue!10}1  \\
 				TempQT~\cite{temqt}     & \cellcolor{Tan!10}-  &\cellcolor{blue!10}1 &\cellcolor{blue!10}1 &/ &\cellcolor{Tan!10}-  \\
 				GlintIQA &\cellcolor{Tan!10}-  &\cellcolor{blue!10}1 & \cellcolor{blue!10}1 &/ &\cellcolor{Tan!10}-  \\
 				\bottomrule
 			\end{tabular}
 		}
 	\end{tabular}
 	\vspace{-3.em}
 \end{table}
\subsubsection{Synthetic database evaluation}  
Table~\ref{sythetic_dataset} summarizes the performance of NR-IQA models on five synthetic distortion datasets. The final two columns present the average performance across four of these datasets, excluding KADID-10k. GlintIQA* was fine-tuned on each dataset using a model pre-trained on our proposed dataset. The proposed GlintIQA* achieves the highest average SROCC and PLCC. Specifically, on the LIVE, most NR-IQA methods perform well, with GlintIQA achieving SROCC and PLCC values close to those of the top-performing DEIQT. For KADID-10k, GlintIQA surpasses DEIQT by 1.91\% in SROCC and 1.80\% in PLCC. GlintIQA attains the highest SROCC on LIVEMD and competitive results on CSIQ. Its performance on TID2013 is slightly lower than that of some methods, as the limited number of original images and the diversity of distortion types introduce variability in the results. However, the pre-trained GlintIQA* shows significant improvements on CSIQ and TID2013, with SROCC and PLCC gains of 3.08\% and 2.85\% on CSIQ, and 7.00\% and 5.21\% on TID2013 compared to GlintIQA. On LIVE, GlintIQA already demonstrates high performance, leaving little room for pre-training improvement. LIVEMD includes compound distortions, while the pre-trained model is based on the proposed dataset with single distortions, resulting in minimal pre-training gains on this dataset.

In addition, the GlintIQA* pre-trained on the proposed dataset outperforms data augmentation methods using FR-IQA annotations (CaHDC and AIGQA), as well as distortion classification-based data augmentation (DB-CNN). This demonstrates the efficacy of the proposed dataset. Furthermore, the proposed dataset can effectively enhance the generalization capability of models, which will be elaborated on in subsequent sections. The proposed dataset is constructed to address the content-limited issue of synthetic distorted datasets. Due to the inherent discrepancy between the distributions of synthetic and authentic distortions \cite{dbcnn}, the pre-trained GlintIQA* model was not evaluated on authentic datasets.

\begin{table}[tbp]
	\centering
	\caption{SROCC evaluation across synthetic distortion datasets.}  
	\fontsize{19pt}{19pt}\selectfont%
	\renewcommand{\arraystretch}{1.1}
	\resizebox{3.5 in}{!}{
		\begin{tabular}{cccccccc}
			\toprule
			\textbf{Training} & \multicolumn{2}{c}{\textbf{LIVE}}& \multicolumn{2}{c}{\textbf{CSIQ}} & \multicolumn{2}{c}{\textbf{TID2013}}&	\multirow{2}{*}{\textbf{Average}}  \\
			\cmidrule(r){2-3} \cmidrule(r){4-5} \cmidrule(r){6-7}
			\textbf{Testing} & \textbf{CSIQ}&\textbf{TID2013}& \textbf{LIVE}&\textbf{TID2013}&\textbf{LIVE}& \textbf{CSIQ}   & \\
			\midrule	
			DIIVINE\cite{diivine}  &0.553&0.487&0.794&0.372 &0.716&0.555&0.580\\
			HOSA\cite{hosa}&0.594& 0.361&0.773 &0.329 &0.846&0.612 &0.586\\
			BRISQUE\cite{brisque}   &0.562 &0.358 &0.847&0.454 &0.774 &0.586&0.597\\
			
			\midrule
			WaDIQaM\cite{WaDIQaM}  & 0.704  &0.462 &-&- &0.817&0.690 & - \\
			HyperIQA \cite{hyperiqa} &0.744 &0.583&0.905 &0.554&0.856&0.692 &0.722\\
			TReS \cite{tres} &0.761 &0.534&0.939 &0.543&0.888&0.708&0.729\\
			DB-CNN \cite{dbcnn}& 0.758 &0.524 &0.877 &0.540&0.891&\textbf{\textcolor{deepgray}{\underline{0.807}}}&0.733\\
			VCRNet~\cite{vcrnet} &0.768 &0.502 & 0.886 &0.542 & 0.822& 0.721&0.707\\
			Hu et al.~\cite{hu_tmm2023} &0.769& 0.584 & 0.924 &0.554 &0.881 &0.736&0.741\\
			\midrule
			\rowcolor{Tan!10}GlintIQA  & \textbf{\textcolor{deepgray}{\underline{0.797}}}&\textbf{\textcolor{deepgray}{\underline{0.664}}} &\textbf{0.945}&\textbf{\textcolor{deepgray}{\underline{0.596}}} &\textbf{\textcolor{deepgray}{\underline{0.936}}}&0.749&\textbf{\textcolor{deepgray}{\underline{0.781}}}\\
			\rowcolor{Tan!10}GlintIQA*& \textbf{0.865}  &\textbf{0.675} &\textbf{\textcolor{deepgray}{\underline{0.944}}}&\textbf{0.695} &\textbf{0.956}&\textbf{0.897} &\textbf{0.839}\\
			\bottomrule
		\end{tabular}
	} 
	\label{cross_syth_dataset} 
\end{table}
\begin{table}[tbp]
	\centering
	\caption{SROCC evaluation across authentic distortion datasets.} 
	\fontsize{20pt}{20pt}\selectfont%
	\renewcommand{\arraystretch}{1.1}
	\resizebox{3.5 in}{!}{ 
		
		\begin{tabular}{cccccccc}
			\toprule
			\textbf{Training} & \multicolumn{2}{c}{\textbf{CLIVE}}& \multicolumn{2}{c}{\textbf{KonIQ-10k}} & \multicolumn{2}{c}{\textbf{BID}} &\multirow{2}{*}{\textbf{Average}} \\
			\cmidrule(r){2-3} \cmidrule(r){4-5} \cmidrule(r){6-7} 
			\textbf{Testing} & \textbf{BID}&\textbf{KonIQ-10k}& \textbf{BID}&\textbf{CLIVE}&\textbf{CLIVE}& \textbf{KonIQ-10k} & \\
			\midrule
			
			DB-CNN \cite{dbcnn}&0.762&0.754&0.816&0.755 &0.725&0.724 & 0.756\\
			HyperIQA \cite{hyperiqa}&0.756&\textbf{\textcolor{deepgray}{\underline{0.772}}}&0.819&0.785 &0.770 &0.688& 0.765\\
			
			REQA \cite{reqa}&0.825&0.762&0.772&\textbf{\textcolor{deepgray}{\underline{0.833}}} &0.747&0.699& 0.773\\
			TReS \cite{tres}&\textbf{\textcolor{deepgray}{\underline{0.859}}}&0.733&0.815&0.786 &0.754&0.707&0.776 \\
			DEIQT\cite{deiqt}&0.852&0.744&\textbf{\textcolor{deepgray}{\underline{0.849}}}&0.794 &\textbf{\textcolor{deepgray}{\underline{0.787}}}&\textbf{\textcolor{deepgray}{\underline{0.731}}}& \textbf{\textcolor{deepgray}{\underline{0.793}}} \\
			
			\midrule
			\rowcolor{Tan!10}GlintIQA &\textbf{0.883}  &\textbf{0.794} &\textbf{0.858}&\textbf{0.859} &\textbf{0.837}&\textbf{0.767}&\textbf{0.833} \\
			\bottomrule
		\end{tabular}
	}	
	\label{cross_auth_dataset}
\end{table}
\subsubsection{Statistical significant test} 
Following the method in \cite{Sharpness_li},  we employed the F-test to assess the statistical significance of the proposed methods in comparison to competing approaches across diverse datasets. Experiments were conducted at a 95\% confidence level to ensure the robustness and reliability of the results. The F-test results are presented in Table~\ref{combined_table}, where '1', '0', and '-' indicate that the model in the row is statistically superior to, inferior to, or indistinguishable from the model in the column, respectively. '/' denotes that performance data is unavailable for that dataset, precluding comparison. For the five authentic distortion datasets, GlintIQA demonstrates statistically significant superiority over DB-CNN, HyperIQA, and TReS. It shows comparable performance to TempQT on the SPAQ dataset while outperforming it on the remaining four datasets. Compared to DEIQT, GlintIQA underperforms on LIVEFB but performs better or comparably on the other datasets. On the five synthetic distortion datasets, GlintIQA* was benchmarked against other methods and exhibited superior performance or was on par with the competitors. Overall, the proposed methods exhibit robust and consistent performance across both types of distortion datasets, underscoring their efficacy in addressing various types of image distortions.

\subsection{Comparison with NR-IQA Methods on Cross-dataset Tests}
\label{Comparisons_cross}
To rigorously evaluate the generalization abilities of NR-IQA models, we trained them on a complete database and subsequently assessed their performance on distinct, unseen databases. To ensure the reliability of results, each experiment was repeated ten times with randomized initializations, and the median of the resulting SROCC scores is presented. The experiments of our model were trained for 300 epochs and tested at 5 epoch intervals. 

\subsubsection{Cross synthetic database evaluation} Table~\ref{cross_syth_dataset} presents results for six cross-synthetic database tests. The proposed GlintIQA and GlintIQA* models demonstrate exceptional performance across various cross-synthetic database tests, where GlintIQA achieves an average SROCC of 0.781, while GlintIQA* achieves an even higher average SROCC of 0.839. In particular, GlintIQA surpasses Hu et al.'s method~\cite{hu_tmm2023} by 3.64\% and 13.70\% when trained on LIVE and tested on CSIQ and TID2013, respectively. When trained on CSIQ, GlintIQA outperforms TReS by 0.64\% on LIVE and the second-ranked method by 7.58\% on TID2013. Trained on TID2013 and tested on LIVE, GlintIQA outperforms DB-CNN by 5.05\%, and DB-CNN outperforms GlintIQA by 7.74\% on CSIQ, attributed to additional data used for pre-training. However, GlintIQA* with extra pre-training data achieves an 11.30\% improvement over DB-CNN. Furthermore, GlintIQA* demonstrates advancements in several other cross-dataset tests, \textit{e.g.}, an 8.53\% improvement over GlintIQA when trained on LIVE and tested on CSIQ, and a remarkable 16.60\% improvement when trained on CSIQ and tested on TID2013. The results highlight the effectiveness of our proposed dataset for enhancing model generalization by providing diverse image content for the model to learn distortion and quality variations.  

\begin{figure}[t!]
	\centering
	\includegraphics[scale=.4]{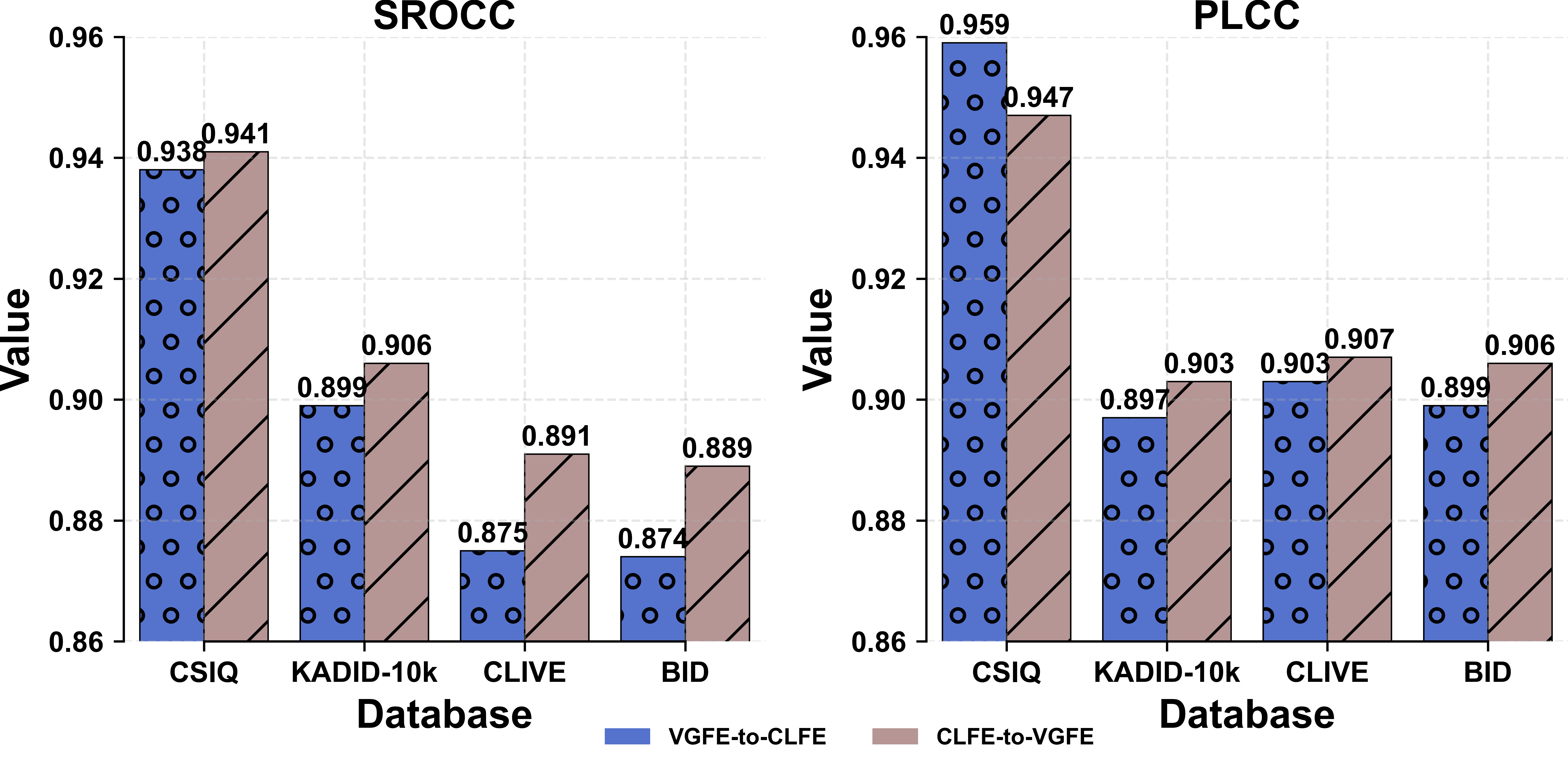} \vspace{-.5em} 	
	\caption{Ablation Study Results on feature integation strategies.}  
	\label{integration_strategy_comparison} 
\end{figure}
\begin{figure}[!tp]
	\centering
	\includegraphics[scale=.495]{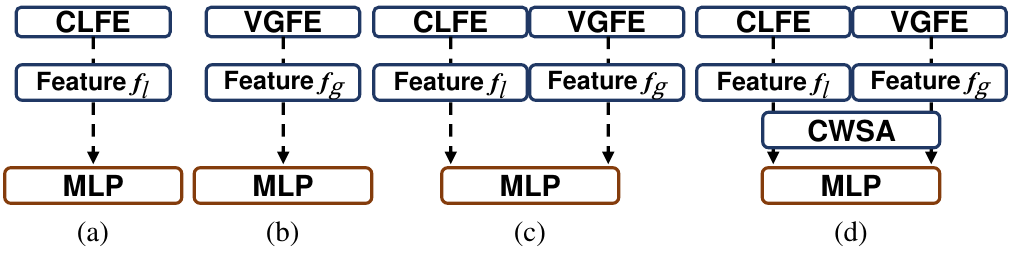} 	\vspace{-1.em}
	\caption{Schematic diagrams of different model structures.}   	%\vspace{-1.em}
	\label{ablation_models} 
\end{figure} 
\subsubsection{Cross authentic database evaluation} Table~\ref{cross_auth_dataset} presents the performance of six cross-authentic dataset tests. The proposed GlintIQA demonstrates the best results across all six cross-authentic database tests. GlintIQA achieves the highest overall average performance, surpassing the second-ranked DEIQT by a significant 5.04\%. When trained on CLIVE, GlintIQA excels on both BID and KonIQ-10k, outperforming TReS by 2.79\% and HyperIQA by 2.85\%, respectively. GlintIQA, when trained on KonIQ-10k, surpasses DEIQT by 1.06\% on BID and REQA by 3.12\% on CLIVE. Notably, GlintIQA achieves even more impressive gains on the relatively small-scale BID dataset, outperforming the second-ranked methods in CLIVE and KonIQ-10k by 6.35\% and 4.92\%, respectively. These exceptional results across both synthetic and authentic database evaluations demonstrate GlintIQA's superior generalization capabilities, attributable to its dual focus on global and local features for capturing image distortions. The combination of convolutional networks and ViT further enhances its ability to learn and generalize effectively across diverse image quality assessment scenarios. 

\begin{table}[tp!]
	\centering
	\caption{Ablation studies of different components of GlintIQA on individual dataset tests.} 
	\fontsize{17.pt}{17.pt}\selectfont%
	\renewcommand{\arraystretch}{1.1}
	\resizebox{3.5 in}{!}{
		\begin{tabular}{lcccccc}
			\toprule
			\multirow{2}{*}{\textbf{Setting}}&	 \multicolumn{2}{c}{\textbf{LIVE}} & \multicolumn{2}{c}{\textbf{CLIVE}} &\multicolumn{2}{c}{\textbf{Average}}\\
			\cmidrule(r){2-3} \cmidrule(r){4-5}  \cmidrule(r){6-7} 
			&\textbf{SROCC} & \textbf{PLCC} & \textbf{SROCC} & \textbf{PLCC}& \textbf{SROCC} & \textbf{PLCC} \tabularnewline
			\midrule
			a) CLFE&0.9629 &0.9661 &0.8289  &0.8323&0.8959~\textbf{\textcolor{Tan}{  ~~~-~~~~~}}&0.8992~\textbf{\textcolor{Tan}{  ~~~-~~~~~}} \tabularnewline
			b) VGFE&0.9720 &0.9753 &0.8810&0.9050 &0.9265~\textbf{\textcolor{Tan}{  +3.42\%}}&0.9402~\textbf{\textcolor{Tan}{  +4.56\%}}  \tabularnewline
			c) CLFE+VGFE &0.9733 &0.9767 &0.8856 &0.9061 &0.9295~\textbf{\textcolor{Tan}{  +3.75\%}}&0.9414~\textbf{\textcolor{Tan}{  +4.69\%}}  \tabularnewline
			d) CLFE+VGFE+CWSA&0.9744 &0.9769&0.8883 &0.9058 &0.9314~\textbf{\textcolor{Tan}{  +3.96\%}} &0.9414~\textbf{\textcolor{Tan}{  +4.69\%}}  \tabularnewline
			
			\rowcolor{Tan!10}e) Proposed&0.9756&0.9797&0.8907& 0.9072&0.9332~\textbf{\textcolor{Tan}{  +4.16\%}}&0.9435~\textbf{\textcolor{Tan}{  +4.93\%}} \tabularnewline
			\bottomrule
		\end{tabular} 
	} 
	\label{ablation_single_dataset} 
\end{table}
\begin{table}[tp!]
	\centering
	\caption{Ablation studies of different components of GlintIQA on cross-dataset tests, where the results are SROCC values.}
	\fontsize{15.pt}{15.pt}\selectfont%
	\renewcommand{\arraystretch}{1.1}
	\resizebox{3.5in}{!}{
		\begin{tabular}{lccccc}
			\toprule
			\multirow{2}{*}{\textbf{Setting}}& \multicolumn{2}{c}{\textbf{LIVE}} & \multicolumn{2}{c}{\textbf{CLIVE}} &\multirow{2}{*}{\textbf{Average}}\\
			\cmidrule(r){2-3} \cmidrule(r){4-5} 
			& \textbf{CSIQ} & \textbf{TID2013} & \textbf{BID} & \textbf{KonIQ-10k}& {} \tabularnewline
			\midrule
			a) CLFE&0.7572  &0.5792&0.8374&0.7306&0.7261~\textbf{\textcolor{Tan}{~~~~~~-~~~~~}} \tabularnewline
			b) VGFE&0.7890  &0.6366&0.8691&0.7687&0.7659~\textbf{\textcolor{Tan}{  +5.48\%}}\tabularnewline
			c) CLFE+VGFE&0.7772  &0.6417&0.8738&0.7798&0.7681~\textbf{\textcolor{Tan}{  +5.78\%}}\tabularnewline
			d) CLFE+VGFE+CWSA&0.7868  &0.6545&0.8780&0.7723&0.7729~\textbf{\textcolor{Tan}{  +6.45\%}}\tabularnewline
			\rowcolor{Tan!10}e) Proposed&0.7973  &0.6637&0.8831&0.7938&0.7845~\textbf{\textcolor{Tan}{  +8.04\%}}\tabularnewline
			\bottomrule
		\end{tabular} 
	} 
	\label{ablation_cross_dataset} 
\end{table}
\subsection{Ablation Study} \label{Ablation}
\subsubsection{Integation strategy} In this ablation study, we explore two feature fusion strategies for integrating multi-scale features from CLFE and VGFE. ``CLFE-to-VGFE": This strategy starts with the initial fusion of CLFE features, followed by a progressive integration with VGFE features. As illustrated in Fig.~\ref{fig:framework}, it emphasizes the importance of global representations while using local details as a supportive element. ``VGFE-to-CLFE": Conversely, this strategy begins with the fusion of VGFE features, followed by a progressive integration of CLFE features. As shown in Fig. \ref{integration_strategy_comparison}, the ``CLFE-to-VGFE" strategy consistently outperforms the ``VGFE-to-CLFE" strategy across most datasets. The results underscore the superior importance of global features in quality assessment. Furthermore, it demonstrates that local details effectively complement VGFE’s global representations, enhancing overall performance. Nevertheless, both strategies achieve relatively high performance, demonstrating their viability in feature integration.

\subsubsection{Model components} To validate the effectiveness of individual components in the proposed GlintIQA, we conduct two types of ablation studies: evaluations on a single dataset and across-dataset tests. The model configurations are illustrated in Fig.~\ref{ablation_models} (omitting feature concatenation, reshaping, and flattening operations for clarity). The structures are described as follows: \textbf{a)}~Utilize CLFE module to extract local multi-scale features and perform regression using MLP on the embedded features. \textbf{b)}~Employ VGFE module to extract global multi-scale features and regress them with MLP. \textbf{c)}~Jointly leverage both CLFE and VGFE modules to extract local and global multi-scale features followed by regression. \textbf{d)}~The proposed model sans the SIEM modules. \textbf{e)}~The model in its complete form, as shown in Fig.~\ref{fig:framework}.
 
Based on the performance comparisons in Tables~\ref{ablation_single_dataset} and~\ref{ablation_cross_dataset}, where the \textbf{\textcolor{Tan}{brown}}-highlighted values indicate performance improvements over model \textbf{a)}, we make the following observations: Model \textbf{b)} with VGFE surpasses the CLFE-based model by 3.42\% in SROCC on individual datasets and 5.48\% on cross-dataset evaluations, demonstrating the superiority of global features in capturing comprehensive quality information. Model \textbf{c)}, combining CLFE and VGFE, shows consistent improvements with 3.75\% and 5.78\% gains in SROCC for individual and cross-dataset evaluations respectively, verifying the complementary nature of local and global features. Model \textbf{d)} with CWSA achieves 3.96\% and 6.45\% improvements in SROCC for respective evaluations, indicating enhanced feature saliency detection. Model \textbf{e)}, integrating both CWSA and SIEM, exhibits optimal performance with 4.16\% and 8.04\% SROCC improvements respectively, demonstrating the effectiveness of their interactive combination in multi-scale feature integration for accurate evaluation of quality assessment.
\begin{figure}[tbp!]
	\centering
	\includegraphics[scale=.26]{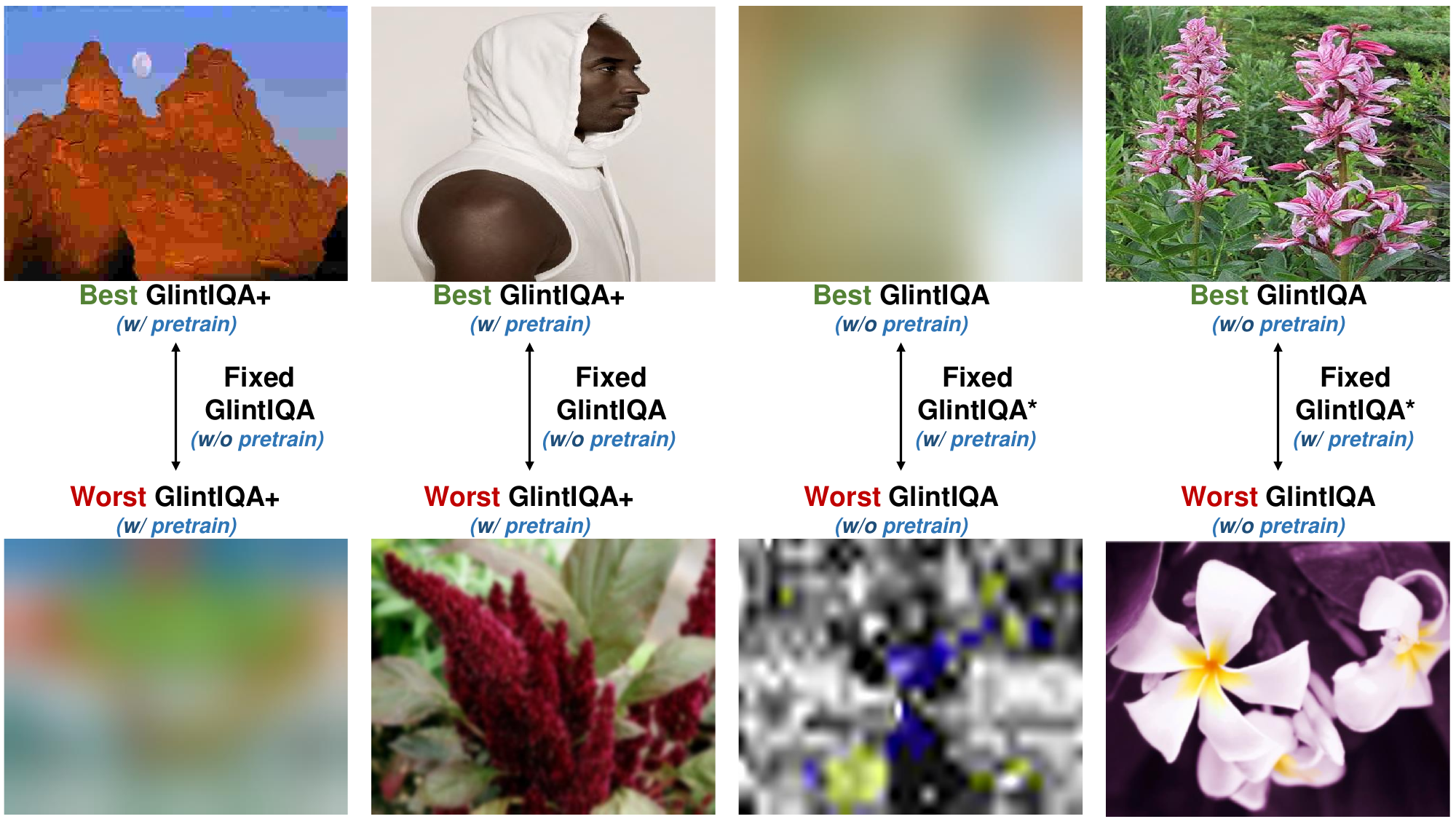} 
	\caption{Results of gMAD competition between GlintIQA and GlintIQA*. The left two columns show fixed GlintIQA at low- and high-quality levels, respectively. The right two columns show fixed GlintIQA* at low- and high-quality levels, respectively.}  
	\label{gmad2} 
\end{figure}
\subsubsection{The proposed SAQT-IQA dataset}~To validate the effectiveness of our proposed dataset, besides the initial validations in Tables~\ref{sythetic_dataset} and~\ref{cross_syth_dataset}, we also conducted cross-dataset tests on KADID-10k, pitting GlintIQA against two other NR-IQA models, \textit{i.e.}, TReS and DEIQT. Their experimental results were replicated utilizing the publicly accessible code under the same training and test sets. As shown in Table~\ref{cross_dataset_kadid}, the results shown that even without pretraining, GlintIQA confidently surpasses both TReS and DEIQT in these evaluations. Furthermore, the pretrained GlintIQA*  amplifies these advantages, boasting average SROCC and PLCC improvements of 5.42\% and 5.72\% over its non-pre-trained counterpart, GlintIQA. Notably,  our proposed dataset is constructed using the labels of KADID-10k. The performance gains achieved by the pre-trained model can be attributed to two key factors: our dataset sufficiently increases the diversity of image content and the effectiveness of the image quality label annotation method.

To facilitate visual comparison, we further conducted a group maximum differentiation (gMAD) competition~\cite{gmad} between GlintIQA and GlintIQA* on Waterloo database~\cite{wateloo}. In the evaluation, a defender model categorizes image quality into six levels, while an attacker model aims to expose its weaknesses by identifying image pairs with significant quality discrepancies within those levels. Specifically, the attacker selects images that the defender model rates similarly but that the attacker model perceives as having a large quality gap. This approach reveals differences in quality perception between models. Fig.~\ref{gmad2} shows the gMAD competition between GlintIQA and pretrained GlintIQA*. When fixed as the defender GlintIQA, GlintIQA* successfully recognizes perceptually different image pairs, whereas the attack by GlintIQA results in image pairs with negligible quality differences. This verifies that pretrained GlintIQA* can enhance model generalizability,  solidifying the validity of our proposed dataset.
\begin{table}[tbp!]
	\centering
	\caption{Ablation study of the proposed SAQT-IQA dataset on cross-database tests and comparison of NR-IQA models.}  
	\scalebox{0.9}[0.9]{
		\renewcommand{\arraystretch}{1.}
		\begin{tabular}{ccccccc}
			\toprule
			\multicolumn{7}{l}{\textbf{\textit{Training on the KADID-10k}}}  \\  
			\hdashline
			\multirow{2}{*}{\textbf{Testing}} & \multicolumn{2}{c}{\textbf{LIVE}}&   \multicolumn{2}{c}{\textbf{CSIQ}} & \multicolumn{2}{c}{\textbf{Average}} \\
			\cmidrule(r){2-3} \cmidrule(r){4-5} \cmidrule(r){6-7}
			\multirow{2}{*}{} & \textbf{SROCC} & \textbf{PLCC} & \textbf{SROCC} & \textbf{PLCC} & \textbf{SROCC} & \textbf{PLCC}\\
			\midrule
			
			DEIQT~\cite{deiqt}&0.9071  &0.8844 & 0.7942 & 0.8021 & 0.8507 & 0.8433 \\
			TReS~\cite{tres}&0.9195&0.9071& 0.7932 & 0.8250 & 0.8563 & 0.8661\\
			
			\midrule
			\rowcolor{Tan!10}GlintIQA&\textbf{\textcolor{deepgray}{\underline{0.9389}}}&\textbf{\textcolor{deepgray}{\underline{0.9158}}}& \textbf{\textcolor{deepgray}{\underline{0.8097}}} & \textbf{\textcolor{deepgray}{\underline{0.8276}}}&\textbf{\textcolor{deepgray}{\underline{ 0.8743}}} & \textbf{\textcolor{deepgray}{\underline{0.8717}}} \\
			\rowcolor{Tan!10}GlintIQA*&\textbf{0.9503}&\textbf{0.9305}&\textbf{0.8931}&\textbf{0.9127}  & \textbf{0.9217} & \textbf{0.9216} \\
			\bottomrule
		\end{tabular}
	} 
	\label{cross_dataset_kadid}
\end{table}

%%%%%%%%%%%%%%%%%%%%%%%%%%%%%%%%%%%%%%%%%%%%%%%%%%%%%%%%%%
\section{Conclusion} \label{conclusion}

This study have addressed key limitations in NR-IQA by focusing on advancements in both model architecture and training data. We proposed  GlintIQA, a novel NR-IQA model that leverages the synergy between CNNs and ViTs to extract complementary local and global multi-grained features. To effectively leverage these features, we proposed a progressive integration strategy that employs cascaded channel-wise self-attention and spatial interaction enhancement modules to dynamically weight feature channels and enhance spatial interactions, leading to improved quality assessment accuracy. To address limited content diversity in synthetic distortion datasets, we developed the semantic-aligned quality transfer method that enables the creation of the SAQT-IQA dataset, This dataset effectively mitigates data scarcity limitations and enhances model generalization. The collective impact of the GlintIQA architecture and the SAQT-IQA dataset is a demonstrably superior NR-IQA performance, confirmed through extensive experimental validation across diverse datasets.

\appendices

\ifCLASSOPTIONcaptionsoff

\fi
\bibliographystyle{IEEEtran}
 
\bibliography{IEEEabrv,reference}
 
\end{document}